\documentclass{article}

\usepackage[preprint]{corl_2026} %

\usepackage{amsmath}
\usepackage{amssymb}
\usepackage{booktabs}
\usepackage{multirow}
\usepackage{graphicx}
\usepackage{placeins}

\title{GeneralVLA-2: Geometry-Aware Reconstruction and Governed Memory for Robot Planning}

\author{
Haoyu Wang$^{1*}$\quad
Guoqing Ma$^{2*}$\quad
Zeyu Zhang$^{1*\dag}$\quad
Yandong Guo$^{3}$\quad
Boxin Shi$^{1}$\quad
Hao Tang$^{1\ddag}$
\\
[0.3em]
$^1$School of Computer Science, Peking University\quad
$^2$CASIA\quad
$^3$AI$^2$ Robotics\\
[0.1em]
\footnotesize $^*$Equal contribution.
$^\dag$Project lead.
$^\ddag$Corresponding author: bjdxtanghao@gmail.com.
}

\begin{document}
\maketitle

\begin{abstract}
Generalist vision-language-action systems need object-centric 3D evidence and reusable manipulation experience to plan reliable robot trajectories. GeneralVLA provides a hierarchical interface for converting language and RGB-D observations into 3D end-effector paths, but two bottlenecks remain. First, monocular SAM3D-style object reconstruction can hallucinate pose and unseen geometry, while manipulation benefits from stable object shape when calibrated multi-view observations are available. Second, the original KnowledgeBank mainly retrieves semantically similar snippets and appends new knowledge, which makes it difficult to control memory quality, conflicts, confidence, and geometric relevance. To address the first challenge, we introduce GeoFuse-MV3D, a geometry-prior-guided MV-SAM3D reconstruction branch that verifies external geometry cues with input-view masks, applies soft visual-hull support, performs axis-wise refinement, and fuses only geometry while preserving appearance. To address the second challenge, we upgrade KnowledgeBank into a governed long-term memory system with explicit quality, confidence, lifecycle, verifier, and conflict metadata, together with precision-oriented retrieval. Finally, we evaluate the reconstruction branch on GSO-30 and the memory module on Terminal-Bench 2.0 and SWE-Bench Verified; GeoFuse-MV3D improves over the MV-SAM3D baseline by reducing CD and LPIPS by 2.20\% and 2.02\% while increasing PSNR and SSIM by 2.36\% and 1.03\%, and KnowledgeBank improves over ReasoningBank by 4.53\% on Terminal-Bench SR and 3.73\% on SWE-Bench resolve rate, while reducing AS by 4.95\% and 5.65\%, respectively.
Code: \url{https://github.com/AIGeeksGroup/GeneralVLA-2}.
Website: \url{https://aigeeksgroup.github.io/GeneralVLA-2}. 
\end{abstract}


\section{Introduction}
\label{sec:introduction}

\paragraph{Background.}
Robotic manipulation increasingly relies on hierarchical vision-language-action (VLA) systems that separate perception, spatial reasoning, and low-level control. GeneralVLA follows this design: a SAM-based affordance segmentation~\citep{kirillov2023segment} and iterative localization module extracts task-relevant visual evidence, 3DAgent reasons over language and depth-projected 3D scene points, and a low-level policy executes the resulting end-effector trajectory~\citep{ma2026generalvla}. This decomposition is attractive because the intermediate 3D trajectory is interpretable and can be corrected with better object geometry or better planning memory. In GeneralVLA-2, we therefore focus on two planner-facing interfaces: an added object-geometry evidence branch for 3DAgent and the KnowledgeBank context retrieved from past manipulation experience.

\paragraph{Challenge.}
The first challenge is reliable object reconstruction from robot observations. Single-image SAM3D-style reconstruction can generate plausible 3D objects from one image and mask, but it is vulnerable to monocular pose ambiguity and hallucinated backside structure~\citep{sam3d2025}. This is problematic for manipulation, where a small pose or shape error can change the grasp, clearance, or collision relation. The second challenge is reliable experience reuse. The original KnowledgeBank stores natural-language memories from past trajectories, but semantic similarity alone does not guarantee that a retrieved memory is safe, current, non-conflicting, or geometrically applicable to the present scene. These two challenges are distinct but coupled: the planner needs both faithful current-scene geometry and trustworthy long-term manipulation knowledge.

\paragraph{Motivation.}
For the first challenge, calibrated RGB-D sensing provides a practical opportunity: when short multi-view observations around an object are available, they can reduce the need for monocular hallucination. The reconstruction update must remain conservative because downstream planning is sensitive to missing parts, over-shrunk geometry, and color or opacity drift. For the second challenge, past manipulation experience should be treated as governed knowledge rather than an append-only text cache. A memory item should carry evidence about where it came from, how reliable it is, whether it is active or stale, what conflicts it has, and whether it matches the current geometry well enough to influence a new trajectory.

\paragraph{Contribution.}
Our first contribution addresses the reconstruction challenge with GeoFuse-MV3D, an improved MV-SAM3D branch for single-object multi-view generation. GeoFuse-MV3D treats external 3D estimation as a geometry-prior provider rather than as the method itself: it validates the prior against input masks, applies appearance calibration, performs soft visual-hull~\citep{laurentini1994visual} and axis-wise geometry refinement, and uses conservative geometry-only fusion while preserving the fixed input views used by the MV-SAM3D benchmark. Our second contribution addresses the experience-reuse challenge by upgrading KnowledgeBank into a governed long-term memory system. The new records store quality, confidence, lifecycle state, verifier metadata, usage statistics, and conflict links; verifier signals control admission and promotion; and retrieval is precision-oriented rather than purely semantic. Our third contribution is an experimental evaluation: we benchmark reconstruction on GSO-30 and evaluate the KnowledgeBank module on Terminal-Bench 2.0 and SWE-Bench Verified.

\paragraph{Summary.}
The resulting GeneralVLA-2 keeps the hierarchical GeneralVLA interface but strengthens the two forms of evidence that condition 3DAgent. GeoFuse-MV3D supplies more stable object geometry under the multi-view reconstruction setting, while governed memory supplies more controlled experience reuse. On GSO-30, GeoFuse-MV3D improves over the MV-SAM3D baseline across CD, PSNR, SSIM, and LPIPS, and ablations show that geometry-prior guidance, mask support verification, and conservative fusion provide complementary gains. On the agent benchmarks, governed memory improves over no-memory and prior memory baselines while reducing AS, supporting the use of explicit memory governance for long-horizon planning.

\section{Related Work}
\label{sec:related-work}

\paragraph{Object-centric 3D reconstruction.}
Single-image 3D methods can produce plausible objects, but they are vulnerable to pose ambiguity and hallucinated unseen structure. SAM3D reconstructs geometry, texture, and layout from one image and mask~\citep{sam3d2025}, while Fast-SAM3D improves the speed-quality tradeoff~\citep{fastsam3d2026}. MV-SAM3D extends this line to multi-view input and adaptive fusion, using complementary observations to improve reconstruction without additional training~\citep{mvsam3d2026}. GeoFuse-MV3D builds on MV-SAM3D, but uses external feed-forward geometry estimates only as priors. In our implementation this prior is instantiated with VGGT~\citep{vggt2025}; the contribution is the mask-verified, appearance-preserving fusion mechanism that makes such priors usable inside the MV-SAM3D reconstruction protocol.

\paragraph{VLA and 3D robot planning.}
Vision-language-action systems connect visual observations, language instructions, and robot actions. RT-2, Octo, OpenVLA, $\pi_0$, and TinyVLA demonstrate the value of large-scale visual-language and robot data for generalist control~\citep{brohan2023rt2,ghosh2024octo,kim2024openvla,black2024pi0,wen2024tinyvla}. A complementary line uses foundation models to produce intermediate plans or 3D representations, as in Code-as-Policies, VoxPoser, and Manipulate-Anything~\citep{DBLP:conf/icra/LiangHXXHIFZ23,DBLP:conf/corl/HuangWZL0023,duan2024manipulateanything}. GeneralVLA follows this interpretable intermediate-representation design: SAM-based affordance segmentation, iterative localization, 3DAgent planning, and low-level execution are separated~\citep{ma2026generalvla}. GeneralVLA-2 keeps this hierarchy and strengthens the two inputs that condition 3DAgent: refined object-centric geometry and retrieved long-term experience.

\paragraph{Memory, verification, and governance.}
Long-horizon agents benefit from reusing past experience, as shown by Synapse, Agent Workflow Memory, and ReasoningBank~\citep{zheng2023synapse,wang2024awm,ouyang2025reasoningbank}. However, semantic relevance alone does not guarantee that a memory is safe, current, non-conflicting, or geometrically applicable to a new manipulation scene. LLM-as-a-judge and verifier-oriented methods provide scalable signals for judging open-ended trajectories, especially when criteria are decomposed and score tokens are handled carefully~\citep{gu2024llmjudge,kwok2026llmverifier}. GeneralVLA-2 uses verifier signals as persistent KnowledgeBank metadata, so retrieval depends not only on textual similarity but also on quality, confidence, lifecycle state, conflict status, and failure-aware constraints.

\section{Preliminaries}
\label{sec:preliminaries}

\paragraph{Task setting.}
We consider a stream of tabletop manipulation tasks. At task $t$, the robot receives a language instruction $q_t$ and an RGB-D observation $o_t=(I_t,D_t)$ from one or more calibrated cameras. The objective is to generate an executable robot trajectory that satisfies the instruction under the current object arrangement, obstacle layout, and grasping constraints. Following GeneralVLA, the system decomposes this problem into three stages: affordance perception, 3D trajectory planning, and low-level execution~\citep{ma2026generalvla}.

\paragraph{Affordance segmentation and 3D point construction.}
The high-level affordance segmentation module (ASM) in GeneralVLA uses SAM-based segmentation~\citep{kirillov2023segment} together with iterative localization to predict task-relevant object and obstacle affordances in the image. Given the depth map and calibrated camera intrinsics, these 2D affordance points are projected into a 3D scene representation $X_t=\{(s_i,P_i^{3D})\}_{i=1}^{n_t}$, where $s_i$ is an object or region label. Multiple 3D points per object help the planner infer extent, pose, opening direction, obstacle height, and other spatial relations that cannot be recovered from a single point.

\paragraph{3DAgent trajectory planning.}
The mid-level 3DAgent receives the instruction $q_t$, the 3D scene representation $X_t$, and a retrieved KnowledgeBank context $B_t$. It produces a coarse end-effector path
\begin{equation}
    \tau_t = \{(x_\ell,y_\ell,z_\ell,g_\ell)\}_{\ell=1}^{L_t},
\end{equation}
where $(x_\ell,y_\ell,z_\ell)$ is a 3D waypoint and $g_\ell$ denotes the gripper state. This trajectory is then passed to the low-level 3D-aware policy and grasping module for execution. The trajectory is therefore an intermediate contract: it must be semantically aligned with the language goal, geometrically valid in the current scene, and easy for the low-level controller to execute.

\paragraph{Original KnowledgeBank.}
GeneralVLA equips 3DAgent with a KnowledgeBank that stores natural-language items extracted from previous trajectories and retrieves semantically similar snippets for future prompts. This gives 3DAgent a useful test-time learning loop, but it leaves three gaps: semantic similarity does not ensure geometric applicability, failed trajectories need to become constraints rather than action recipes, and a growing memory needs lifecycle control to suppress duplicate, conflicting, stale, or low-confidence records.

\section{Method}
\label{sec:method}

\begin{figure}[!t]
    \centering
    \includegraphics[width=\linewidth]{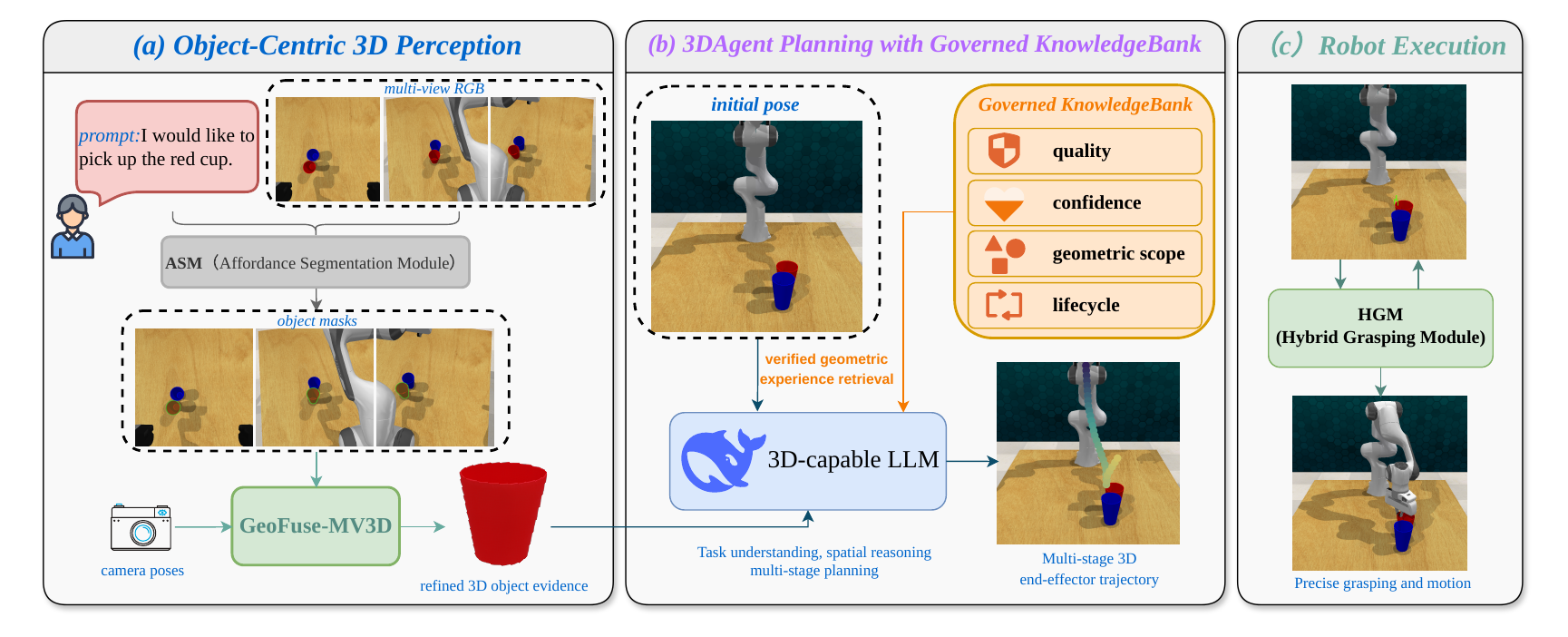}
    \caption{Overview of GeneralVLA-2 for robot manipulation. When calibrated multi-view object observations are available, GeoFuse-MV3D converts the views, masks, and camera poses into refined object-centric 3D evidence; the 3D-capable planning agent combines this evidence with governed KnowledgeBank retrieval; and the robot execution module follows the resulting multi-stage end-effector trajectory.}
    \label{fig:generalvla2-overview}
\end{figure}

Figure~\ref{fig:generalvla2-overview} summarizes the full GeneralVLA-2 pipeline. The system builds object-centric 3D evidence from calibrated observations, using GeoFuse-MV3D when multi-view object inputs are available, then conditions a 3D-capable planning agent on both the refined object geometry and the governed KnowledgeBank, and finally executes the planned trajectory through the robot grasping and motion module. The following subsections describe the two planner-facing improvements: GeoFuse-MV3D for object reconstruction and governed KnowledgeBank for experience reuse.

\FloatBarrier

\subsection{GeoFuse-MV3D Reconstruction Branch}
\label{sec:mvsam3d}

GeneralVLA-2 first adds an object-side multi-view reconstruction branch to complement the original SAM-based affordance segmentation and iterative localization interface. With calibrated multi-view RGB-D observations around the same object, monocular hallucination becomes less necessary than it is in a single-image setup. We build on MV-SAM3D and introduce GeoFuse-MV3D as a conservative multi-view reconstruction branch: it uses the fixed input views and masks to regularize geometry, and it applies small continuous corrections rather than hard deletion.

\begin{figure}[!htbp]
    \centering
    \includegraphics[width=\linewidth]{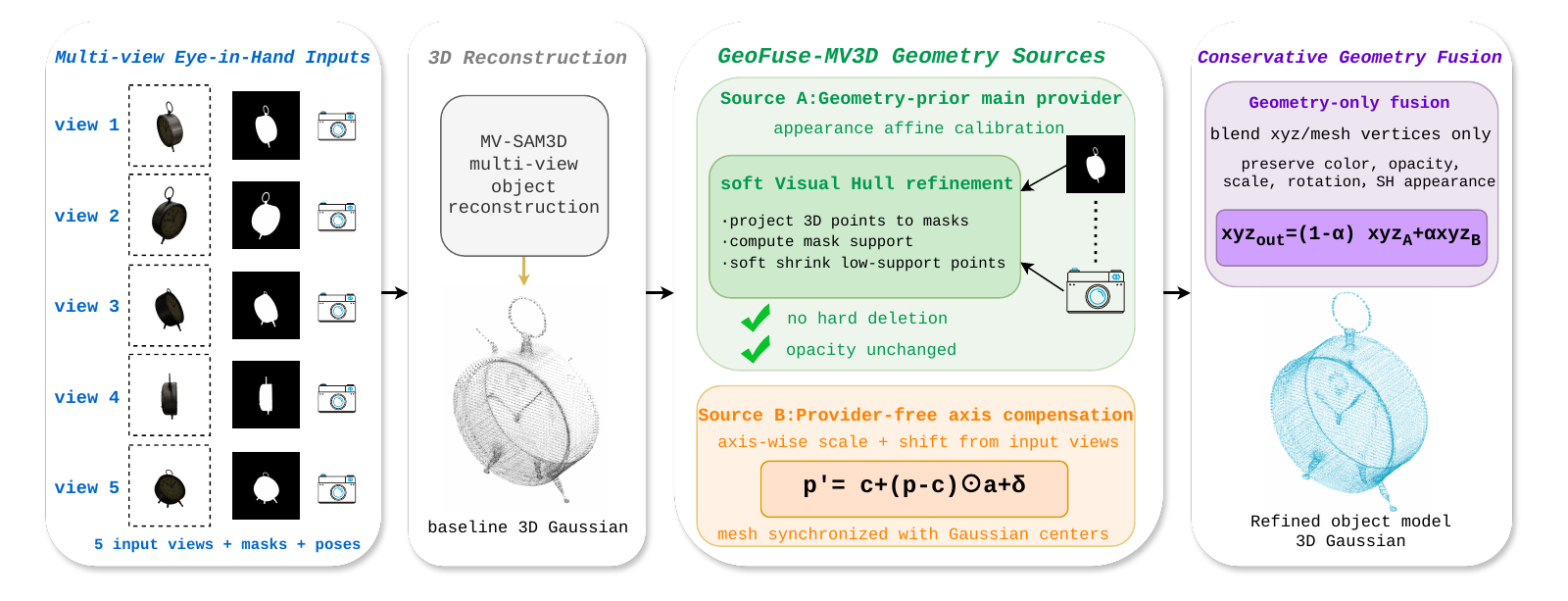}
    \caption{GeoFuse-MV3D reconstruction branch in GeneralVLA-2. The branch keeps the same multi-view inputs, masks, and poses as MV-SAM3D, then refines the baseline with two complementary geometry sources and conservative geometry-only fusion.}
    \label{fig:geofuse-mv3d}
\end{figure}
\FloatBarrier

As shown in Figure~\ref{fig:geofuse-mv3d}, GeoFuse-MV3D is organized around two geometry sources. Let the multi-view input be
\begin{equation}
    \mathcal{D}
    = \{(I_i,M_i,K_i,T_i)\}_{i \in \mathcal{V}_{\mathrm{in}}},
    \qquad
    \mathcal{V}_{\mathrm{in}}=\{0,1,2,3,4\},
\end{equation}
where $I_i$ is the RGB observation, $M_i$ is the object mask, $K_i$ is the camera intrinsic matrix, and $T_i$ is the camera pose. MV-SAM3D produces an initial Gaussian object
\begin{equation}
    G_0=\{(x_0^j,\theta_0^j)\}_{j=1}^{N},
\end{equation}
where $x_0^j \in \mathbb{R}^3$ is a Gaussian center~\citep{kerbl20233dgaussians} and $\theta_0^j$ contains the non-geometric rendering attributes, including opacity, scale, rotation, and spherical-harmonic appearance. Source A starts from the MV-SAM3D output, incorporates an external geometry-prior provider instantiated with VGGT in our implementation, and applies lightweight appearance affine calibration. Source B is an input-view axis-compensation branch that does not use the external provider.

Both geometry sources are checked against the same input masks. For a point $p$, let $\mathcal{V}(p)$ be the input views where it projects inside the image, and let $M_i(\pi_i(p))$ be the bilinearly sampled mask value. The mask-consistency score is
\begin{equation}
    s(p) = \frac{1}{\max(|\mathcal{V}(p)|,1)} \sum_{i \in \mathcal{V}(p)} M_i(\pi_i(p)).
\end{equation}
Instead of deleting low-support points or reducing opacity, GeoFuse-MV3D converts low support into a small inward geometry correction:
\begin{equation}
    p' = c + (p-c)\,(1-\lambda(p)),
\end{equation}
with $c$ the object center and $\lambda(p)$ bounded by a small maximum shrink ratio. Source A also applies a lightweight appearance affine calibration, but the final fusion remains geometry-only so that color, opacity, scale, rotation, and SH appearance fields are not overwritten.

We then apply a low-dimensional axis-wise correction to the Gaussian centers and synchronize the same transform to the mesh vertices. Writing $a \in \mathbb{R}^3$ for axis scales and $\delta \in \mathbb{R}^3$ for a small shift, the corrected point is
\begin{equation}
    p'' = c + (p'-c) \odot a + \delta.
\end{equation}
The axis parameters are selected to improve mask agreement while remaining close to the original MV-SAM3D geometry. This keeps Source B provider-free: it uses only the input masks, poses, and the existing reconstruction, so it is orthogonal to the external geometry-prior branch.

Finally, when the two outputs preserve compatible Gaussian indexing, we blend only the geometry coordinates with
\begin{equation}
    p_{\mathrm{out}} = (1-\alpha)p_A + \alpha p_B,
\end{equation}
while leaving color, opacity, scale, rotation, and SH features unchanged. More generally, when source A provides the trusted non-geometric Gaussian attributes, the final output is
\begin{equation}
    G_{\mathrm{out}}
    =\{((1-\alpha)x_A^j+\alpha x_B^j,\theta_A^j)\}_{j=1}^{N}.
\end{equation}
This keeps the final branch conservative: it can absorb compatible geometry corrections, but it falls back to the trusted source-A result whenever topology differs or mask support is weak. Appendix~\ref{app:geofuse-details} gives the expanded equations for projection, soft visual hull, appearance calibration, and confidence-weighted residual fusion.

\subsection{Governed KnowledgeBank Overview}
\label{sec:method-overview}

\begin{figure}[!t]
    \centering
    \includegraphics[width=\linewidth]{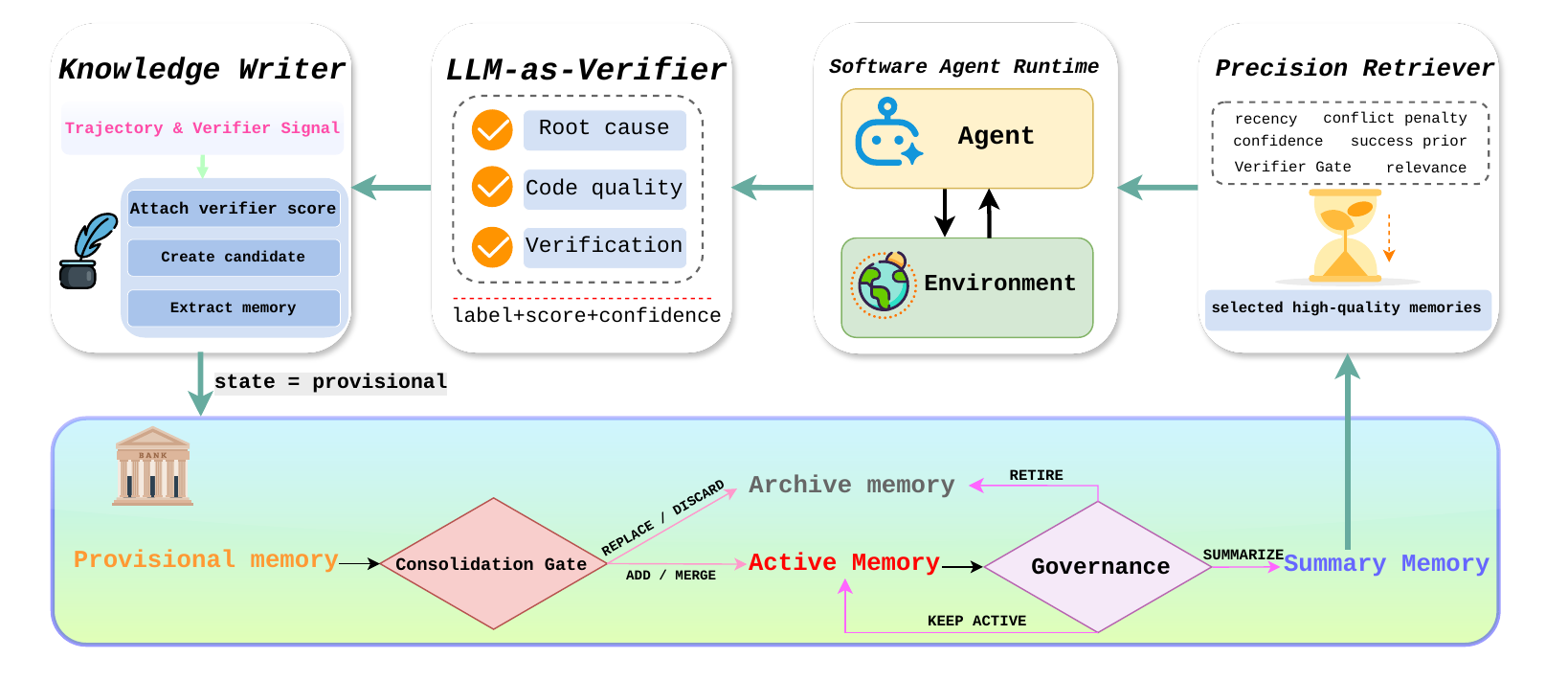}
    \caption{Architecture of the governed KnowledgeBank module used by GeneralVLA-2. The module writes verifier-labeled memories, retrieves high-quality records, and manages their lifecycle before conditioning the 3DAgent planner.}
    \label{fig:knowledgebank-architecture}
\end{figure}

With this object-side evidence, GeneralVLA-2 supplies refined 3D information to 3DAgent when it is available. It also improves the mid-level planner by replacing the original append-oriented KnowledgeBank with a governed KnowledgeBank. Figure~\ref{fig:knowledgebank-architecture} summarizes this memory subsystem. The affordance segmentation interface and low-level execution interface remain unchanged. At each robot task, GeneralVLA-2 retrieves a compact set of reliable memory records, renders them as a structured planning context, and asks 3DAgent to produce a 3D waypoint path. After execution, the completed trajectory, scene representation, final outcome, and execution feedback are verified. The system then extracts candidate knowledge, assigns quality and confidence metadata, and updates the KnowledgeBank through admission, merging, conflict handling, summarization, and archival.

This design follows a simple principle: the robot should not trust either geometry or memory blindly. GeoFuse-MV3D reduces monocular pose hallucination in the multi-view reconstruction setting, while governed memory decides which past manipulation knowledge has enough support to influence the next trajectory. When no sufficiently reliable knowledge exists, the retriever can abstain and 3DAgent plans from the current observation alone.

\subsection{Governed Memory Operations}
\label{sec:governed-memory}

KnowledgeBank stores each reusable manipulation lesson as a structured record rather than as an untyped text snippet:
\begin{equation}
    m = (q,c,y,z,\kappa,R,\mathcal{L},v),
\end{equation}
where $q$ is the source query, $c$ is the reusable content, $y$ is the memory type, $z$ is the lifecycle state, $\kappa$ is confidence, $R$ is a verifier-derived quality score, $\mathcal{L}$ stores conflict or supersession links, and $v$ stores verifier metadata. Memory types separate procedural hints, failure-avoidance constraints, and tool-usage guidance; lifecycle states separate provisional, active, summary, and archived records. This distinction is important for robot planning because a failed trajectory can provide a useful constraint, but should not be replayed as a positive action recipe.

After task $t$, the runtime logs the instruction, scene representation, planned trajectory, execution feedback, and outcome. A verifier scores the candidate knowledge using task completion, spatial consistency, collision safety, execution validity, and generalizability criteria. Following verifier-style score aggregation~\citep{kwok2026llmverifier}, we compute a quality score $R_t$ from criterion-level scores and promote only sufficiently supported candidates. Retrieval is precision-oriented:
\begin{equation}
\begin{split}
    S(q_t,X_t,m) ={}&
    r_{\mathrm{text}}(q_t,m) + \kappa_m + b_{\mathrm{success}}(m)
    + b_{\mathrm{recency}}(m) \\
    &+ b_{\mathrm{usage}}(m) - p_{\mathrm{conflict}}(m)-p_{\mathrm{stale}}(m).
\end{split}
\end{equation}
The retrieved records are rendered as bounded planning context for 3DAgent: procedural memories are optional hints, failure memories are constraints, and stale or conflicting records are suppressed. Consolidation then performs add, merge, replace, discard, summarization, and archival operations under a fixed active-memory budget. Appendix~\ref{app:knowledgebank-details} provides the full record schema and lifecycle details.

\subsection{Integration with 3DAgent}
\label{sec:integration}

GeneralVLA-2 keeps the GeneralVLA hierarchy and strengthens the two interfaces that condition 3DAgent. The SAM-based ASM and iterative localization still provide task-relevant visual affordances, depth projection still defines the planner-facing 3D point interface, and the low-level policy still consumes the waypoint-and-gripper trajectory. GeoFuse-MV3D adds refined object-centric geometry when calibrated multi-view object inputs are available, while the governed KnowledgeBank enriches the retrieved planning context. At inference time, 3DAgent receives the instruction, the current 3D scene representation, the refined object evidence when available, and the structured KnowledgeBank block. It then outputs the same trajectory format expected by the low-level policy. This separation keeps GeneralVLA-2 compatible with the original robot stack while improving the geometry and memory evidence used for planning.

\section{Experiments}
\label{sec:experiments}

\subsection{GeoFuse-MV3D GSO-30 Evaluation}
\label{sec:mvsam3d-experiments}

We first evaluate GeoFuse-MV3D on the GSO-30 single-object multi-view benchmark, derived from Google Scanned Objects~\citep{downs2022google}, following the official MV-SAM3D protocol. For fairness, the MV-SAM3D baseline and all GeoFuse-MV3D variants use the identical GSO-30 object list, the identical five input views $(0,1,2,3,4)$, the identical object masks, and the identical camera poses. We evaluate held-out rendering on target views $(10,\ldots,24)$ and report Chamfer Distance (CD), PSNR, SSIM~\citep{wang2004ssim}, and LPIPS~\citep{zhang2018lpips}. Lower CD and LPIPS are better; higher PSNR and SSIM are better. The baseline numbers are produced by our local reproduction using the official MV-SAM3D code and reported configuration, while GeoFuse-MV3D changes only the reconstruction refinement stage through geometry-prior guidance, mask verification, and geometry-only fusion. The full quantitative table is reported in Appendix~\ref{app:geofuse-evaluation}. In summary, GeoFuse-MV3D improves all four reconstruction metrics relative to the MV-SAM3D baseline: CD and LPIPS decrease by 2.20\% and 2.02\%, while PSNR and SSIM increase by 2.36\% and 1.03\%.
\FloatBarrier

\subsection{KnowledgeBank Evaluation}
\label{sec:knowledgebank-evaluation}

We evaluate the KnowledgeBank memory-governance module outside the robot stack, using Terminal-Bench 2.0~\citep{terminalbench2026} and SWE-Bench Verified~\citep{jimenez2024swebench,openai2024swebenchverified} as controlled long-horizon agent benchmarks. This module-level evaluation isolates whether admission, verification, retrieval, and lifecycle management improve experience reuse before the same governed KnowledgeBank is attached to 3DAgent. Terminal-Bench is reported with success rate (SR) and AS, while SWE-Bench Verified is reported with resolve rate and AS; AS denotes average steps, so lower values indicate more efficient problem solving.

Full results are moved to Appendix~\ref{app:knowledgebank-results}. Across four model backbones, KnowledgeBank improves over ReasoningBank by 4.53\% on Terminal-Bench SR and 3.73\% on SWE-Bench resolve rate, while reducing AS by 4.95\% and 5.65\%, respectively. The result supports the main design choice: reusable experience is more reliable when memory admission, retrieval, conflict handling, and lifecycle updates are explicitly governed rather than treated as append-only semantic retrieval.
\FloatBarrier

\subsection{Training-free Performance for Robotic Planning}
\label{sec:training-free-performance}

\begin{table*}[!htbp]
\setlength{\tabcolsep}{4pt}
\caption{\textbf{Task-averaged success rate \% for training-free evaluation.} Methods are evaluated in simulation tasks from RLBench~\cite{DBLP:journals/ral/JamesMAD20}. 
w/o KnowledgeBank removes retrieved KnowledgeBank guidance from trajectory planning.
Each task was evaluated over 3 seeds to obtain the task-averaged success rate and standard deviations.}
\centering
    \resizebox{1.\textwidth}{!}{
    \begin{tabular}{cccc cccc} %
      \toprule  %
      \enspace \textbf{Method}   & Put\_block & Play\_jenga & Open\_jar &Close\_box & Open\_box & Pickup\_cup &Push\_block  \\
      \midrule
      VoxPoser~\cite{DBLP:conf/corl/HuangWZL0023} &   70.70$\pm$2.31 & 0.00$\pm$0.00  & 0.00$\pm$0.00 & 0.00$\pm$0.00 & 0.00$\pm$0.00 & 26.70$\pm$14.00  & \textbf{25.33}$\pm$8.33 
\\
      CAP~\cite{DBLP:conf/icra/LiangHXXHIFZ23} &  84.00$\pm$16.00 & 0.00$\pm$0.00 & 0.00$\pm$0.00 & 0.00$\pm$0.00 & 0.00$\pm$0.00 &  14.67$\pm$4.62 & 8.00$\pm$4.00
\\
      Hamster~\cite{DBLP:conf/iclr/0038DZJMG0F00025} &  78.33$\pm$6.11  & 0.00$\pm$0.00 & 77.67$\pm$11.55 & 0.00$\pm$0.00 & 0.00$\pm$0.00 & 9.00$\pm$2.26  & 5.00$\pm$6.11
\\
      GeneralVLA-2 (Ours) &  \textbf{90.33}$\pm$8.72  & \textbf{85.33}$\pm$14.05 & \textbf{85.00}$\pm$6.93 & \textbf{54.67}$\pm$12.00 & \textbf{38.33}$\pm$12.86 & \textbf{87.33}$\pm$6.11 & 25.00$\pm$15.53
\\
      GeneralVLA-2 w/o KnowledgeBank &  75.00$\pm$14.05  & 63.33$\pm$11.37 & 68.67$\pm$6.43 & 31.00$\pm$15.53 & 10.00$\pm$12.86 & 76.67$\pm$11.37 & 15.00$\pm$4.00
\\
      \bottomrule %
      \toprule
        \enspace \textbf{Method}   & Take\_umbrella & Sort\_mustard & Open\_wine & Lamp\_on & Put\_knife & Pick\_\&\_lift & Insert\_block  \\
      \midrule
      VoxPoser~\cite{DBLP:conf/corl/HuangWZL0023} &   33.33$\pm$8.33 & \textbf{96.00}$\pm$6.93  & 8.00$\pm$4.00 & 57.30$\pm$12.22 &  \textbf{92.00}$\pm$4.00 & 96.00$\pm$0.00  & 0.00$\pm$0.00
\\
      CAP~\cite{DBLP:conf/icra/LiangHXXHIFZ23} &  4.00$\pm$4.00 & 0.00$\pm$0.00 & 0.00$\pm$0.00 & 64.00$\pm$6.93 & 14.67$\pm$8.33 & \textbf{ 100.00}$\pm$0.00  & 0.00$\pm$0.00
\\
      Hamster~\cite{DBLP:conf/iclr/0038DZJMG0F00025} & 8.67$\pm$2.31  & 44.33$\pm$12.86 & 34.33$\pm$20.13 & 61.00$\pm$8.00 & 23.00$\pm$0.00 & 96.00$\pm$0.00  & 0.00$\pm$0.00
\\
      GeneralVLA-2 (Ours) &  \textbf{68.00}$\pm$15.62  & 79.33$\pm$19.22 & \textbf{44.67}$\pm$14.05 & \textbf{78.67}$\pm$10.58 & 63.67$\pm$12.86 & 90.67$\pm$12.00  & \textbf{34.33}$\pm$11.14
\\
      GeneralVLA-2 w/o KnowledgeBank &  48.00$\pm$15.53  & 57.33$\pm$6.43 & 26.33$\pm$14.05 & 58.67$\pm$15.62 & 43.67$\pm$6.11 & 66.00$\pm$19.22 & 13.33$\pm$6.93
\\
      \bottomrule %
    \end{tabular}
    }
    \label{table:training-free}
    \vspace{6pt} 
\end{table*}

\noindent \textbf{Implementation details.}
We use GeoFuse-MV3D to extract object-centric 3D evidence when multi-view object observations are available, and use DeepSeek-R1~\citep{deepseekai2025deepseekr1} for reasoning over text-serialized 3D points. We evaluate 14 RLBench~\cite{DBLP:journals/ral/JamesMAD20} simulation tasks with a Franka Panda robot, RGB-D observations, and motion-planner execution~\cite{DBLP:journals/ram/SucanMK12}. Before testing, the agent explores each task ten times and stores reusable experience in the KnowledgeBank without parameter training~\cite{intelligence2504pi0}. We compare with CAP~\cite{DBLP:conf/icra/LiangHXXHIFZ23}, Hamster~\cite{DBLP:conf/iclr/0038DZJMG0F00025}, and VoxPoser~\cite{DBLP:conf/corl/HuangWZL0023}; Appendix~\ref{app:robot-evaluation-details} provides the full environment and baseline details.

GeneralVLA-2 generates successful trajectories for all 14 tasks, while Hamster, VoxPoser, and CAP cover only 10, 9, and 7 tasks, respectively (Table~\ref{table:training-free}). It outperforms the baselines in 10 tasks, and removing KnowledgeBank consistently lowers success, showing that governed experience reuse helps trajectory planning. Appendix~\ref{app:robot-evaluation-details} discusses the remaining difficult task categories.

We further ablate GeoFuse-MV3D in Appendix~\ref{app:mvsam3d-ablation}. The ablation shows that the geometry-prior branch is strongest for CD, the provider-free axis branch gives the strongest single-source appearance metrics, and their conservative geometry fusion gives the best PSNR, SSIM, and LPIPS while keeping CD clearly better than the MV-SAM3D baseline.
\FloatBarrier

\subsection{Real-World Robot Experiments}
\label{sec:realworld-experiments}

\textbf{Environment and tasks.} We test GeneralVLA-2 on an Agilex-2.0 Piper manipulator with a parallel gripper and a top-facing Intel RealSense L515 RGB-D camera. The four language-conditioned tasks are move\_spray\_bottle, open\_drawer, open\_jar, and sort\_object, each evaluated over 10 episodes with varying object poses across three trials.

\begin{figure}[!htbp]
  \centering
  \includegraphics[width=.68\textwidth]{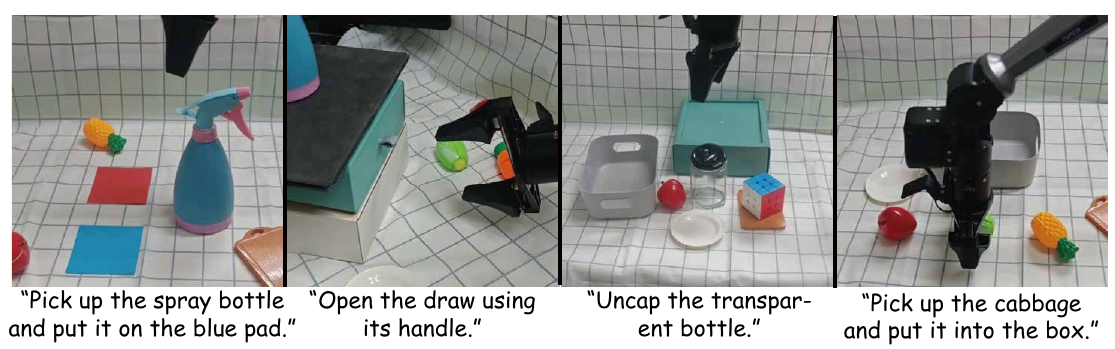}
  \caption{Real-world training-free demonstrations for four manipulation tasks.}
  \label{fig:realtask}
\end{figure}

\textbf{Results.} GeneralVLA-2 succeeds on all four real-world tasks and outperforms CAP and RoboPoint~\citep{DBLP:conf/corl/YuanDBPKMMF24}. Table~\ref{table:realworld} summarizes the quantitative results, and Figure~\ref{fig:realtask} shows representative executions. Appendix~\ref{app:robot-evaluation-details} gives additional task-level interpretation.

\begin{table}[!htbp]
\setlength{\tabcolsep}{1.5pt}
\caption{Success rates for real-world training-free task completion.} %
\centering
\vspace{-5pt} 
    \resizebox{.8\textwidth}{!}{
    \begin{tabular}{ccccc} %
      \toprule  %
      \enspace \textbf{Method}   & Move\_spray\_bottle &  Open\_drawer & Open\_jar & Sort\_object     \\
      \midrule
      CAP (0-shot) &  6.67  & 0.00 & 36.67  & 70.00
\\
      RoboPoint~\citep{DBLP:conf/corl/YuanDBPKMMF24} (0-shot) &  0.00  & 0.00 &  20.00  & 63.33 
\\
      GeneralVLA-2 (training-free) &  \textbf{63.33} & \textbf{40.00} & \textbf{53.33}  & \textbf{83.33}
\\
      \bottomrule %
    \end{tabular}
    }
    \label{table:realworld}
    \vspace{-0.4cm} 
\end{table}

\FloatBarrier

\section{Limitations}
\label{sec:limitations}

GeneralVLA-2 still depends on reliable calibrated observations, masks, poses, and verifier quality; failures in these inputs can propagate to reconstruction or memory retrieval. In the multi-view reconstruction setting, calibration or mask failures can directly affect GeoFuse-MV3D. The method is intentionally conservative, so its gains over MV-SAM3D are consistent but modest, and our real-world evaluation does not yet cover long-horizon mobile manipulation, heavy occlusion, deformable objects, or human-in-the-loop recovery. Appendix~\ref{app:limitations-details} provides additional discussion.

\section{Conclusion}
\label{sec:conclusion}

We presented GeneralVLA-2 as a planner-facing extension of GeneralVLA. GeoFuse-MV3D improves object-centric 3D evidence under calibrated multi-view inputs through mask-verified geometry support, provider-agnostic priors, axis-wise refinement, and appearance-preserving fusion. The governed KnowledgeBank improves experience reuse through verifier-aware records with quality, confidence, lifecycle, conflict, and usage metadata. Experiments show that GeoFuse-MV3D improves GSO-30 reconstruction over the MV-SAM3D baseline under the same input-view protocol, while KnowledgeBank improves Terminal-Bench 2.0 and SWE-Bench Verified results across multiple backbones. Together, these results suggest that hierarchical VLA systems benefit from improving the planner inputs directly: more faithful object geometry and more controlled long-term memory.

\bibliography{verified_refs}  %

\clearpage
\appendix
\section*{Appendix}
\section{Additional GeoFuse-MV3D Evaluation}
\label{app:geofuse-evaluation}

This appendix collects the complete GeoFuse-MV3D reconstruction evidence: the main GSO-30 quantitative result, additional qualitative comparisons, implementation equations, and component ablations. All GeoFuse-MV3D experiments use the same GSO-30 objects, five input views, masks, camera poses, and held-out target views as the reproduced MV-SAM3D baseline.

\subsection{GSO-30 Reconstruction Results}

\begin{table}[!htbp]
    \centering
    \caption{GSO-30 reconstruction results for GeoFuse-MV3D.}
    \label{tab:mvsam3d-gso30}
    \small
    \setlength{\tabcolsep}{6pt}
    \renewcommand{\arraystretch}{1.0}
    \resizebox{\linewidth}{!}{%
    \begin{tabular}{lcccc}
        \toprule
        Method & CD $\downarrow$ ($10^{-3}$) & PSNR $\uparrow$ & SSIM $\uparrow$ & LPIPS $\downarrow$ \\
        \midrule
        MV-SAM3D baseline & 45.8876 & 13.2421 & 0.8051 & 0.2795 \\
        \textbf{GeoFuse-MV3D} & \textbf{44.8770 {\scriptsize(-2.20\%)}} & \textbf{13.5547 {\scriptsize(+2.36\%)}} & \textbf{0.8134 {\scriptsize(+1.03\%)}} & \textbf{0.2739 {\scriptsize(-2.02\%)}} \\
        \bottomrule
    \end{tabular}%
    }
\end{table}

\subsection{Qualitative Reconstruction Results}
\label{app:qualitative-reconstruction}

\begin{figure*}[t]
    \centering
    \makebox[\textwidth][c]{\includegraphics[page=1,width=1.16\textwidth]{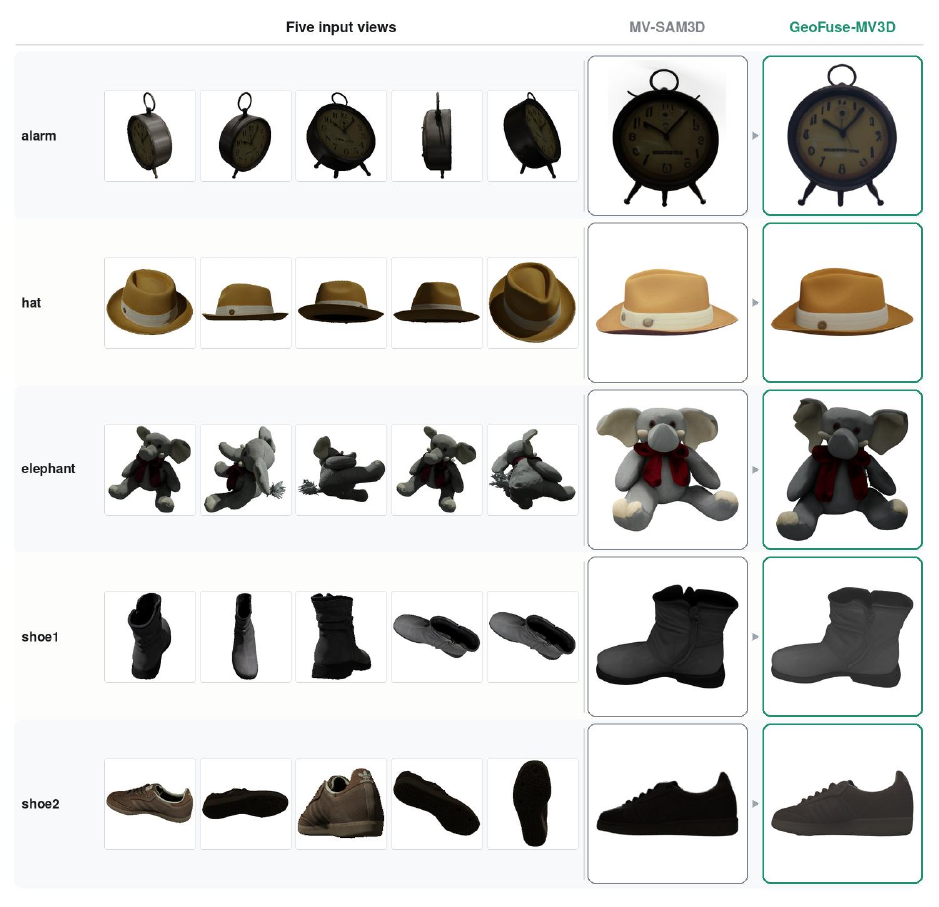}}
    \caption{Additional qualitative comparison between MV-SAM3D and GeoFuse-MV3D on the first set of GSO-30 objects. Each row uses the same five input views, masks, and camera poses for both methods.}
    \label{fig:appendix-mvsam3d-geofuse-a}
\end{figure*}

\begin{figure*}[p]
    \centering
    \makebox[\textwidth][c]{\includegraphics[page=2,width=1.16\textwidth]{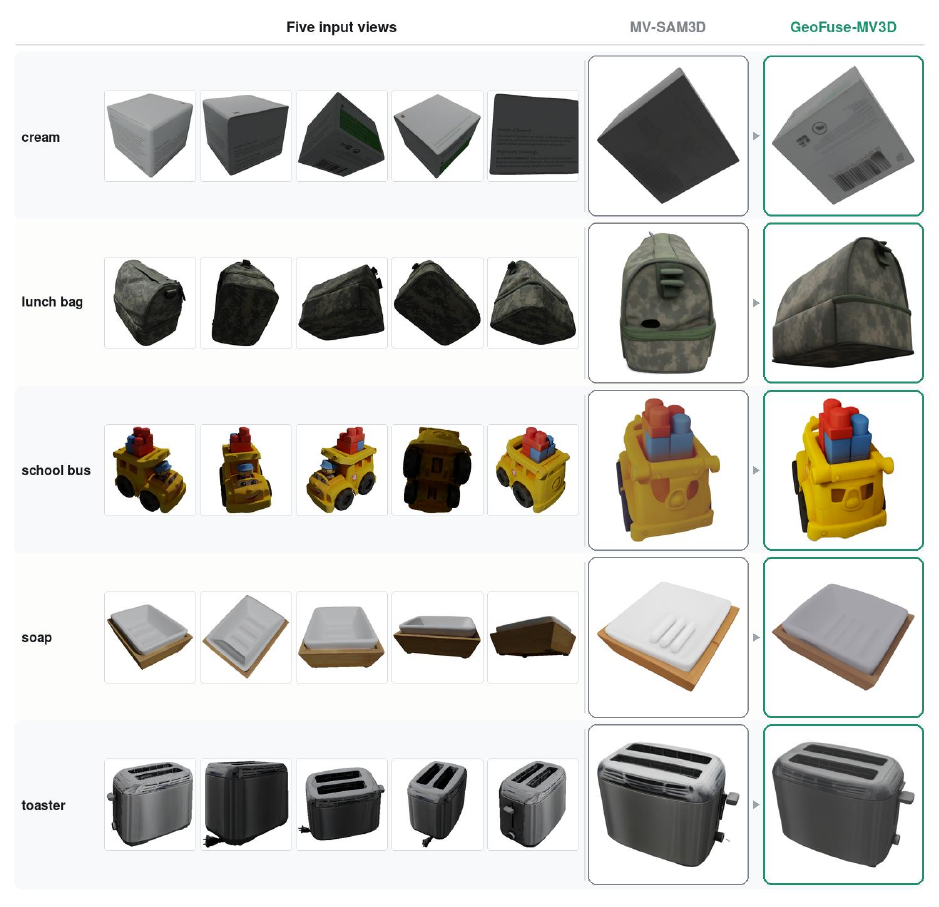}}
    \caption{Additional qualitative comparison between MV-SAM3D and GeoFuse-MV3D on the second set of GSO-30 objects. GeoFuse-MV3D applies mask-verified, appearance-preserving geometry refinement, improving object completeness and pose consistency without changing the input protocol.}
    \label{fig:appendix-mvsam3d-geofuse-b}
\end{figure*}

Figures~\ref{fig:appendix-mvsam3d-geofuse-a} and~\ref{fig:appendix-mvsam3d-geofuse-b} provide additional qualitative comparisons between the MV-SAM3D baseline and GeoFuse-MV3D on GSO-30 objects. For each object, both methods use the same five input views, object masks, and camera poses. This appendix complements the quantitative CD, PSNR, SSIM, and LPIPS results in Section~\ref{sec:mvsam3d-experiments}: GeoFuse-MV3D keeps the evaluation protocol fixed while producing more complete and geometrically consistent object reconstructions in several challenging cases.

\subsection{Additional GeoFuse-MV3D Details}
\label{app:geofuse-details}

This appendix expands the compact GeoFuse-MV3D equations used in Section~\ref{sec:mvsam3d}. Given a 3D point $p$, its projection into input view $i$ is
\begin{equation}
    \pi_i(p) =
    \Pi\!\left(K_i T_i
    \begin{bmatrix}
    p \\ 1
    \end{bmatrix}\right),
\end{equation}
where $\Pi(\cdot)$ denotes perspective division. For a Gaussian set $G$, the average mask reprojection disagreement is
\begin{equation}
    E_{\mathrm{mask}}(G)
    = \frac{1}{N}\sum_{j=1}^{N}\left(1-s(x^j)\right).
\end{equation}
GeoFuse-MV3D uses this as a conservative diagnostic rather than a hard deletion rule. Low support is converted into a bounded shrink strength
\begin{equation}
    \lambda(p) = \lambda_{\max} \cdot \sigma\!\left(\frac{\tau - s(p)}{\eta}\right)^2,
\end{equation}
where $\tau$ is the support threshold, $\eta$ is the softness, and $\lambda_{\max}$ is the maximum shrink ratio.

For Source A, appearance affine calibration estimates a small per-channel gain $g_i$ and bias $b_i$ on masked pixels $\Omega_i=\{u:M_i(u)>0\}$:
\begin{equation}
    (g_i^\star,b_i^\star)
    =
    \arg\min_{g_i,b_i}
    \sum_{u \in \Omega_i}
    \left\|
    g_i \odot \hat{I}_i(u) + b_i - I_i(u)
    \right\|_1
    + \beta\|g_i-\mathbf{1}\|_2^2
    + \beta\|b_i\|_2^2 .
\end{equation}
The regularizers keep calibration close to identity, which helps preserve the input object's original tone.

For Source B, axis compensation chooses scale and shift parameters by
\begin{equation}
    (a^\star,\delta^\star)
    =
    \arg\min_{a,\delta}
    E_{\mathrm{mask}}\!\left(\mathcal{T}_{a,\delta}(G)\right)
    + \rho_a\|a-\mathbf{1}\|_2^2
    + \rho_\delta\|\delta\|_2^2,
\end{equation}
where $\mathcal{T}_{a,\delta}(p)=c+(p-c)\odot a+\delta$. When support scores are available for both sources, geometry fusion can be written as confidence-weighted residual fusion:
\begin{equation}
    x_{\mathrm{out}}^j
    =
    x_A^j
    +
    \alpha\,w^j(x_B^j-x_A^j),
    \qquad
    w^j=\mathrm{clip}\!\left(s(x_A^j)s(x_B^j),0,1\right).
\end{equation}
The non-geometric attributes remain $\theta_{\mathrm{out}}^j=\theta_A^j$.
\FloatBarrier

\subsection{GeoFuse-MV3D Component Ablation}
\label{app:mvsam3d-ablation}

We ablate GeoFuse-MV3D to separate the effect of geometry-prior guidance, input-mask soft visual hull support, the provider-free axis-compensation branch, and multi-source geometry fusion. All rows use the same GSO-30 object set, the same five input views $(0,1,2,3,4)$, and the same held-out target views $(10,\ldots,24)$.

\begin{table}[!htbp]
    \centering
    \caption{Component ablation of GeoFuse-MV3D on GSO-30 full30, averaged over three seeds.}
    \label{tab:mvsam3d-ablation}
    \small
    \setlength{\tabcolsep}{4.6pt}
    \renewcommand{\arraystretch}{1.0}
    \resizebox{\linewidth}{!}{%
    \begin{tabular}{lcccc}
        \toprule
        Variant & CD $\downarrow$ ($10^{-3}$) & PSNR $\uparrow$ & SSIM $\uparrow$ & LPIPS $\downarrow$ \\
        \midrule
        MV-SAM3D baseline & 45.8876 & 13.2421 & 0.8051 & 0.2795 \\
        A: geometry prior + appaff & 45.2204 {\scriptsize(-1.45\%)} & 13.4546 {\scriptsize(+1.60\%)} & 0.8103 {\scriptsize(+0.65\%)} & 0.2876 {\scriptsize(+2.90\%)} \\
        A + softVH & 45.1879 {\scriptsize(-1.52\%)} & 13.4470 {\scriptsize(+1.55\%)} & 0.8104 {\scriptsize(+0.66\%)} & 0.2877 {\scriptsize(+2.93\%)} \\
        B: provider-free axis compensation & 45.7427 {\scriptsize(-0.32\%)} & 13.4985 {\scriptsize(+1.94\%)} & 0.8109 {\scriptsize(+0.72\%)} & 0.2758 {\scriptsize(-1.32\%)} \\
        A+B geometry fusion & \textbf{44.9530} {\scriptsize(-2.04\%)} & \textbf{13.5549} {\scriptsize(+2.36\%)} & \textbf{0.8135} {\scriptsize(+1.04\%)} & \textbf{0.2735} {\scriptsize(-2.15\%)} \\
        \bottomrule
    \end{tabular}%
    }
\end{table}
\FloatBarrier

Table~\ref{tab:mvsam3d-ablation} shows that the geometry-prior main-provider branch is the strongest single source for CD, while the provider-free axis branch gives the strongest single-source PSNR, SSIM, and LPIPS gains. Adding softVH further improves CD and SSIM over source A, but it slightly lowers PSNR and LPIPS because the mask-support constraint prioritizes geometric support over appearance fidelity. Combining the two sources gives the best PSNR, SSIM, and LPIPS among the ablated component rows while keeping CD clearly better than the baseline.
\FloatBarrier

\section{Additional KnowledgeBank Evaluation}
\label{app:knowledgebank-evaluation}

\subsection{Additional KnowledgeBank Details}
\label{app:knowledgebank-details}

The full KnowledgeBank record additionally includes source status, usage statistics, deduplication keys, and verifier metadata:
\begin{equation}
    m = (i, q, c, y, s, z, \kappa, R, u, d, \mathcal{L}, v).
\end{equation}
The lifecycle state $z$ is one of \textit{provisional}, \textit{active}, \textit{summary}, or \textit{archive}. The memory type $y$ is one of \textit{procedural\_hint}, \textit{failure\_avoidance}, or \textit{tool\_usage}. Procedural hints capture reusable positive strategies, failure-avoidance memories record negative lessons from failed or unsafe trajectories, and tool-usage memories describe reusable interaction knowledge for the planning or execution interface.

After a task finishes, the verifier evaluates whether trajectory-derived knowledge should influence future planning. For robot tasks, the criteria correspond to task completion, spatial consistency, collision safety, execution validity, and generalizability. For the module-level software-agent evaluation, the same KnowledgeBank implementation uses root-cause analysis, code quality, and empirical verification criteria. With criterion set $\mathcal{C}$ and score-token set $\mathcal{V}$, the verifier-derived quality score can be written as
\begin{equation}
    R_t =
    \frac{1}{|\mathcal{C}|}
    \sum_{c \in \mathcal{C}}
    \sum_{v \in \mathcal{V}}
    p_\theta(v \mid q_t,X_t,\tau_t,c)\phi(v),
\end{equation}
where $\phi(\cdot)$ maps valid score tokens to $[0,1]$. If log probabilities are unavailable, the system parses the tagged score and falls back to a midpoint score only when no valid token is found. The aggregate score becomes one of \textit{verified\_success}, \textit{verified\_fail}, or \textit{uncertain}.

Consolidation compares a candidate with active records using query relevance, content relevance, signature similarity, deduplication keys, and conflict checks. The edit action is one of \textit{ADD}, \textit{MERGE}, \textit{REPLACE}, or \textit{DISCARD}. Governance periodically retires stale failed records, resolves conflicts within each memory type, creates summary memories from repeated successful clusters, and archives overflow when the active budget is exceeded.

\subsection{KnowledgeBank Benchmark Results}
\label{app:knowledgebank-results}

Table~\ref{tab:knowledgebank-results} reports the full module-level KnowledgeBank evaluation on Terminal-Bench 2.0 and SWE-Bench Verified. The backbone models are Qwen-3.5-Flash, Qwen-3.5-Plus~\citep{qwen2026qwen35}, Gemini-3-Flash, and Gemini-3.1-Pro~\citep{googledeepmind2025gemini3flash,googledeepmind2026gemini31pro}. All metrics report mean $\pm$ standard deviation over five seeds. The experiment compares no memory, prior memory baselines, and the governed KnowledgeBank under the same backbone models and benchmark protocols.

\begin{table}[!htbp]
    \centering
    \caption{KnowledgeBank results on Terminal-Bench 2.0 and SWE-Bench Verified.}
    \label{tab:knowledgebank-results}
    \scriptsize
    \setlength{\tabcolsep}{4.2pt}
    \renewcommand{\arraystretch}{0.86}
    \resizebox{\linewidth}{!}{%
    \begin{tabular}{llcccc}
        \toprule
        \multirow{2}{*}{Model} & \multirow{2}{*}{Memory Method}
        & \multicolumn{2}{c}{Terminal-Bench 2.0}
        & \multicolumn{2}{c}{SWE-Bench Verified} \\
        \cmidrule(lr){3-4} \cmidrule(lr){5-6}
        & & SR $\uparrow$ & AS $\downarrow$
        & Resolve Rate $\uparrow$ & AS $\downarrow$ \\
        \midrule
        \multirow{4}{*}{Qwen-3.5-Flash}
        & No Memory     & $48.5{\pm}1.6$ & $42.8{\pm}2.7$ & $67.0{\pm}1.1$ & $45.6{\pm}1.4$ \\
        & AWM           & $50.1{\pm}1.3$ & $41.8{\pm}2.2$ & $68.4{\pm}0.8$ & $44.2{\pm}1.6$ \\
        & ReasoningBank & $52.8{\pm}2.0$ & $40.9{\pm}1.5$ & $70.8{\pm}1.5$ & $42.5{\pm}1.1$ \\
        & \textbf{KnowledgeBank} & \textbf{$55.8{\pm}1.8$} & \textbf{$38.8{\pm}2.5$} & \textbf{$73.4{\pm}1.2$} & \textbf{$39.9{\pm}1.0$} \\
        \midrule
        \multirow{4}{*}{Qwen-3.5-Plus}
        & No Memory     & $60.8{\pm}1.0$ & $34.2{\pm}1.9$ & $76.6{\pm}1.3$ & $35.0{\pm}0.8$ \\
        & AWM           & $60.6{\pm}1.7$ & $34.6{\pm}2.4$ & $78.2{\pm}0.7$ & $34.3{\pm}1.6$ \\
        & ReasoningBank & $64.6{\pm}2.1$ & $32.1{\pm}1.4$ & $80.1{\pm}0.9$ & $32.7{\pm}1.5$ \\
        & \textbf{KnowledgeBank} & \textbf{$67.4{\pm}1.9$} & \textbf{$30.4{\pm}1.7$} & \textbf{$83.6{\pm}1.6$} & \textbf{$30.6{\pm}1.2$} \\
        \midrule
        \multirow{4}{*}{Gemini-3-Flash}
        & No Memory     & $55.6{\pm}1.5$ & $37.4{\pm}2.6$ & $74.2{\pm}0.8$ & $40.7{\pm}1.4$ \\
        & AWM           & $56.8{\pm}1.6$ & $36.4{\pm}2.9$ & $74.6{\pm}1.0$ & $40.6{\pm}0.9$ \\
        & ReasoningBank & $59.0{\pm}1.2$ & $35.6{\pm}1.1$ & $78.0{\pm}1.5$ & $37.9{\pm}1.3$ \\
        & \textbf{KnowledgeBank} & \textbf{$61.8{\pm}1.7$} & \textbf{$34.1{\pm}2.3$} & \textbf{$80.4{\pm}1.4$} & \textbf{$36.1{\pm}1.0$} \\
        \midrule
        \multirow{4}{*}{Gemini-3.1-Pro}
        & No Memory     & $68.4{\pm}0.8$ & $30.7{\pm}2.7$ & $78.2{\pm}1.3$ & $28.5{\pm}1.2$ \\
        & AWM           & $70.1{\pm}2.1$ & $30.1{\pm}2.0$ & $79.9{\pm}1.1$ & $28.0{\pm}0.8$ \\
        & ReasoningBank & $73.0{\pm}1.0$ & $28.8{\pm}1.3$ & $82.2{\pm}1.6$ & $26.8{\pm}1.7$ \\
        & \textbf{KnowledgeBank} & \textbf{$75.7{\pm}1.3$} & \textbf{$27.3{\pm}2.5$} & \textbf{$85.3{\pm}1.2$} & \textbf{$25.4{\pm}1.4$} \\
        \bottomrule
    \end{tabular}%
    }
\end{table}
\FloatBarrier

KnowledgeBank outperforms the no-memory and prior-memory baselines across both benchmarks and all model backbones. ReasoningBank is already a strong memory baseline: compared with No Memory, it improves Terminal-Bench SR by 4.0 points and SWE-Bench resolve rate by 3.8 points on average, while also reducing AS. The governed KnowledgeBank further improves over ReasoningBank, showing that memory quality control, conflict handling, and lifecycle management improve experience reuse beyond semantic retrieval alone.

Table~\ref{tab:overhead} reports the corresponding deployment cost. KnowledgeBank adds verifier and governance calls, which appear in the extra-token column, but the shorter executions reduce total tokens and latency relative to AWM and ReasoningBank for every backbone. No Memory remains the lowest-cost reference because it removes retrieval and verification, but it also loses the accuracy gains reported in Table~\ref{tab:knowledgebank-results}.

\begin{table}[!htbp]
\centering
\scriptsize
\setlength{\tabcolsep}{2.4pt}
\renewcommand{\arraystretch}{0.88}
\caption{Deployment cost accounting for memory-augmented agent execution.}
\label{tab:overhead}
\resizebox{\linewidth}{!}{%
\begin{tabular}{llcccccc}
\toprule
{\textbf{Model}} & {\textbf{Method}} & {\textbf{AS}} & {\textbf{Agent tok./task}} & {\textbf{Extra tok./task}} & {\textbf{Total tok./task}} & {\textbf{Latency/task}} & {\textbf{Storage MB}} \\
\midrule
\multirow{4}{*}{\textbf{\textit{Qwen-3.5-Flash}}}
& No Memory & 29.8 & 64.0k & 0.0k & 64.0k & 103.2s & 0.0 \\
& AWM & 28.9 & 67.0k & 0.0k & 67.0k & 112.8s & 1.6 \\
& ReasoningBank & 27.8 & 68.4k & 0.0k & 68.4k & 115.4s & 2.5 \\
& \textbf{KnowledgeBank} & \textbf{25.3} & \textbf{61.0k} & \textbf{4.0k} & \textbf{65.0k} & \textbf{108.9s} & \textbf{3.6} \\
\midrule
\multirow{4}{*}{\textbf{\textit{Qwen-3.5-Plus}}}
& No Memory & 23.9 & 78.5k & 0.0k & 78.5k & 164.2s & 0.0 \\
& AWM & 23.5 & 83.0k & 0.0k & 83.0k & 179.3s & 1.5 \\
& ReasoningBank & 22.0 & 84.0k & 0.0k & 84.0k & 184.0s & 2.1 \\
& \textbf{KnowledgeBank} & \textbf{19.9} & \textbf{76.5k} & \textbf{4.5k} & \textbf{81.0k} & \textbf{174.8s} & \textbf{3.0} \\
\midrule
\multirow{4}{*}{\textbf{\textit{Gemini-3-Flash}}}
& No Memory & 26.5 & 71.5k & 0.0k & 71.5k & 101.8s & 0.0 \\
& AWM & 25.9 & 76.4k & 0.0k & 76.4k & 110.8s & 1.5 \\
& ReasoningBank & 24.6 & 77.0k & 0.0k & 77.0k & 114.0s & 2.3 \\
& \textbf{KnowledgeBank} & \textbf{22.7} & \textbf{67.9k} & \textbf{4.9k} & \textbf{72.8k} & \textbf{106.9s} & \textbf{3.3} \\
\midrule
\multirow{4}{*}{\textbf{\textit{Gemini-3.1-Pro}}}
& No Memory & 20.6 & 130.5k & 0.0k & 130.5k & 329.0s & 0.0 \\
& AWM & 20.0 & 140.0k & 0.0k & 140.0k & 357.0s & 1.4 \\
& ReasoningBank & 19.0 & 141.2k & 0.0k & 141.2k & 363.0s & 2.0 \\
& \textbf{KnowledgeBank} & \textbf{17.3} & \textbf{127.0k} & \textbf{5.8k} & \textbf{132.8k} & \textbf{344.0s} & \textbf{2.9} \\
\bottomrule
\end{tabular}%
}
\vspace{2pt}
\begin{minipage}{\linewidth}
\raggedright
\scriptsize\emph{Note.} AS is the unweighted macro-average over the agent benchmark settings used for cost accounting. Agent tokens include task execution and compact memory-injection prompts; extra tokens count verifier, re-verification, induction, and governance calls outside AS. No Memory is a no-retrieval, no-verification lower-cost reference rather than an accuracy--cost optimum. Storage is the final serialized memory-bank footprint.
\end{minipage}
\end{table}
\FloatBarrier

The main cost tradeoff is therefore not simply extra verifier usage; it is whether verifier outputs are stored as persistent metadata that can improve future retrieval and reduce the number of agent steps.

\subsection{KnowledgeBank Component Ablation}
\label{app:knowledgebank-ablation-section}

We ablate the major components of the governed KnowledgeBank to isolate the source of the gains. The ablations remove the admission module (w/o Adm.), replay module (w/o Rep.), governance module (w/o Gov.), and failure handling module (w/o Fail.). We also evaluate a semantic-retrieval-only variant (Sem. Ret.) that retrieves memory by textual similarity without the full governance pipeline.

\begin{table}[!htbp]
    \centering
    \caption{Ablation study of KnowledgeBank.}
    \label{tab:knowledgebank-ablation}
    \scriptsize
    \setlength{\tabcolsep}{4.4pt}
    \renewcommand{\arraystretch}{0.84}
    \resizebox{\linewidth}{!}{%
    \begin{tabular}{llcccc}
        \toprule
        \multirow{2}{*}{Model} & \multirow{2}{*}{Variant}
        & \multicolumn{2}{c}{Terminal-Bench 2.0}
        & \multicolumn{2}{c}{SWE-Bench Verified} \\
        \cmidrule(lr){3-4} \cmidrule(lr){5-6}
        & & SR $\uparrow$ & AS $\downarrow$
        & Resolve Rate $\uparrow$ & AS $\downarrow$ \\
        \midrule
        \multirow{6}{*}{Qwen-3.5-Flash}
        & \textbf{Full} & \textbf{55.8} & \textbf{38.8} & \textbf{73.4} & \textbf{39.9} \\
        & w/o Adm.  & 52.7 & 40.5 & 70.5 & 41.7 \\
        & w/o Rep.  & 53.9 & 39.9 & 71.2 & 41.2 \\
        & w/o Gov.  & 51.6 & 41.4 & 69.1 & 43.0 \\
        & w/o Fail. & 54.8 & 39.8 & 71.8 & 41.1 \\
        & Sem. Ret. & 51.5 & 41.5 & 68.8 & 43.1 \\
        \midrule
        \multirow{6}{*}{Qwen-3.5-Plus}
        & \textbf{Full} & \textbf{67.4} & \textbf{30.4} & \textbf{83.6} & \textbf{30.6} \\
        & w/o Adm.  & 64.4 & 32.0 & 80.9 & 32.5 \\
        & w/o Rep.  & 65.3 & 31.5 & 81.5 & 31.7 \\
        & w/o Gov.  & 63.7 & 32.6 & 79.8 & 33.0 \\
        & w/o Fail. & 66.1 & 31.4 & 82.2 & 31.5 \\
        & Sem. Ret. & 63.4 & 32.9 & 79.4 & 33.2 \\
        \midrule
        \multirow{6}{*}{Gemini-3-Flash}
        & \textbf{Full} & \textbf{61.8} & \textbf{34.1} & \textbf{80.4} & \textbf{36.1} \\
        & w/o Adm.  & 58.9 & 35.8 & 78.1 & 37.7 \\
        & w/o Rep.  & 59.6 & 35.3 & 78.4 & 37.2 \\
        & w/o Gov.  & 57.6 & 36.5 & 75.9 & 38.8 \\
        & w/o Fail. & 60.6 & 35.0 & 78.9 & 37.1 \\
        & Sem. Ret. & 57.4 & 36.6 & 75.2 & 39.0 \\
        \midrule
        \multirow{6}{*}{Gemini-3.1-Pro}
        & \textbf{Full} & \textbf{75.7} & \textbf{27.3} & \textbf{85.3} & \textbf{25.4} \\
        & w/o Adm.  & 72.1 & 29.1 & 81.5 & 27.0 \\
        & w/o Rep.  & 73.8 & 28.2 & 83.3 & 26.4 \\
        & w/o Gov.  & 71.0 & 29.4 & 79.8 & 27.9 \\
        & w/o Fail. & 74.3 & 28.0 & 83.8 & 26.3 \\
        & Sem. Ret. & 70.7 & 29.6 & 79.0 & 28.2 \\
        \bottomrule
    \end{tabular}%
    }
\end{table}
\FloatBarrier

Table~\ref{tab:knowledgebank-ablation} shows that each component contributes to the complete memory module. Removing governance causes one of the largest drops, reducing Terminal-Bench SR by 4.2 points and SWE-Bench resolve rate by 4.4 points on average, while increasing AS on both benchmarks. The Sem. Ret. variant is weaker still, dropping 4.4 Terminal-Bench SR points and 5.0 SWE-Bench resolve-rate points relative to the full system. These results support the central design choice of GeneralVLA-2's KnowledgeBank: high-value memory requires lifecycle management, conflict control, and failure-aware filtering rather than semantic retrieval alone. The smaller but consistent degradation from removing failure handling further suggests that failed trajectories are useful only when they are explicitly converted into constraints instead of being replayed as ordinary skill priors.
\FloatBarrier

\section{Additional Robot Evaluation Details}
\label{app:robot-evaluation-details}

For the simulation experiments in Section~\ref{sec:training-free-performance}, the environment uses a Franka Panda robot with a parallel gripper, CoppeliaSim, and PyRep. Four RGB-D cameras capture tabletop observations. The 14 RLBench tasks cover different object categories, object poses, and task horizons, and robot actions are represented as waypoints executed by a motion planner. CAP uses language models to generate programs that call hand-crafted primitive actions; VoxPoser predicts waypoints through 3D voxel value maps; and Hamster uses a VLM to generate 2D end-effector trajectories. We provide CAP with ground-truth simulation states and object models, and VoxPoser with segmented object point clouds, making the comparison conservative for GeneralVLA-2.

The remaining difficult simulation tasks are mainly non-prehensile or fine-grained manipulation tasks that require more precise 3D pose estimation or dynamic correction during execution. VoxPoser is limited in tasks that require moving the arm beyond 4-DoF, while CAP is constrained by its hand-written primitive set. The gap between GeneralVLA-2 and GeneralVLA-2 without KnowledgeBank indicates that retrieved manipulation experience supplies useful task-level constraints beyond the current observation alone.

\begin{figure}[!htbp]
    \centering
    \scriptsize
    \setlength{\tabcolsep}{2pt}
    \renewcommand{\arraystretch}{0.92}
    \begin{tabular}{ccccc}
        \toprule
        Task & Start & Early & Mid & Final \\
        \midrule
        \raisebox{0.085\linewidth}{Buttons} &
        \includegraphics[width=0.19\linewidth]{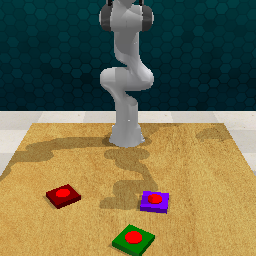} &
        \includegraphics[width=0.19\linewidth]{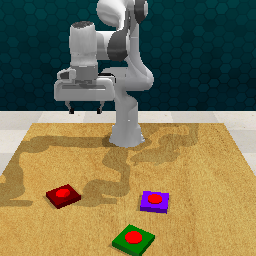} &
        \includegraphics[width=0.19\linewidth]{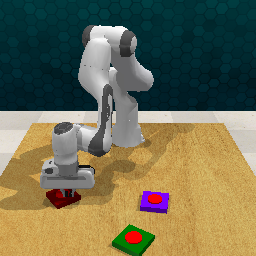} &
        \includegraphics[width=0.19\linewidth]{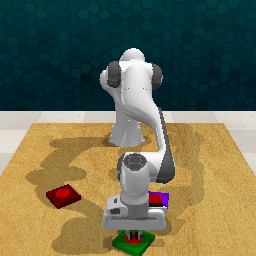} \\
        \raisebox{0.085\linewidth}{Cup} &
        \includegraphics[width=0.19\linewidth]{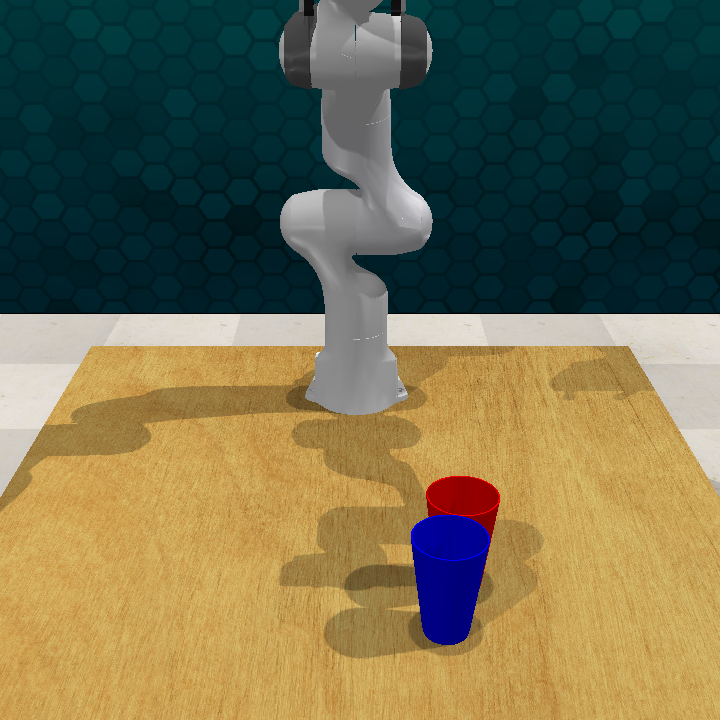} &
        \includegraphics[width=0.19\linewidth]{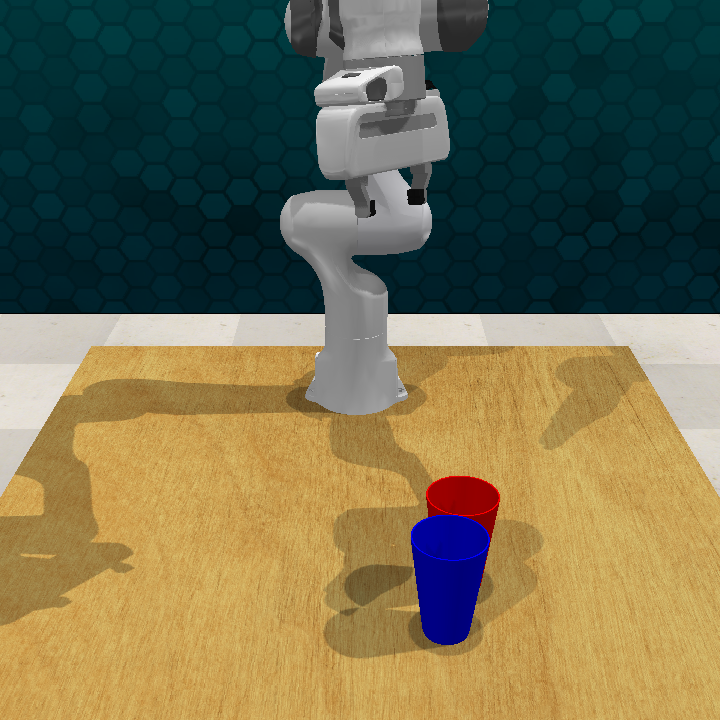} &
        \includegraphics[width=0.19\linewidth]{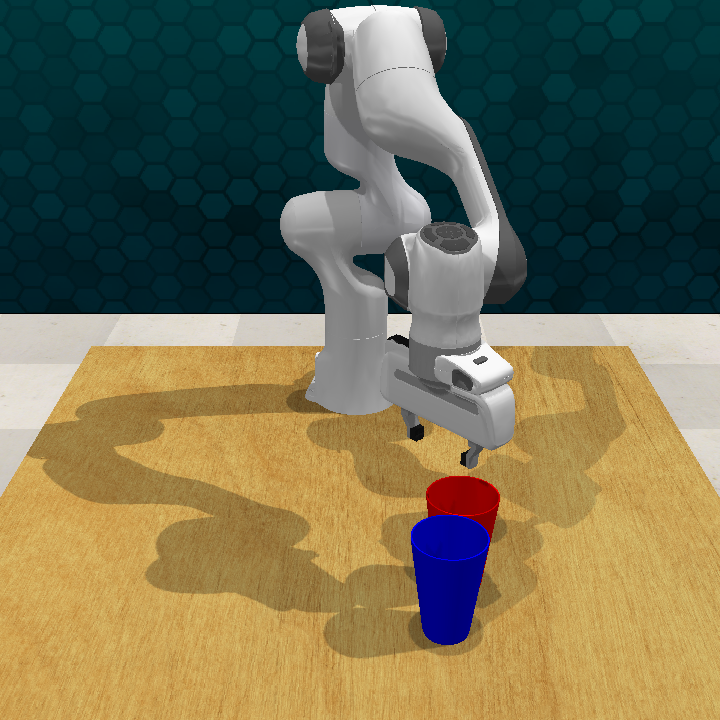} &
        \includegraphics[width=0.19\linewidth]{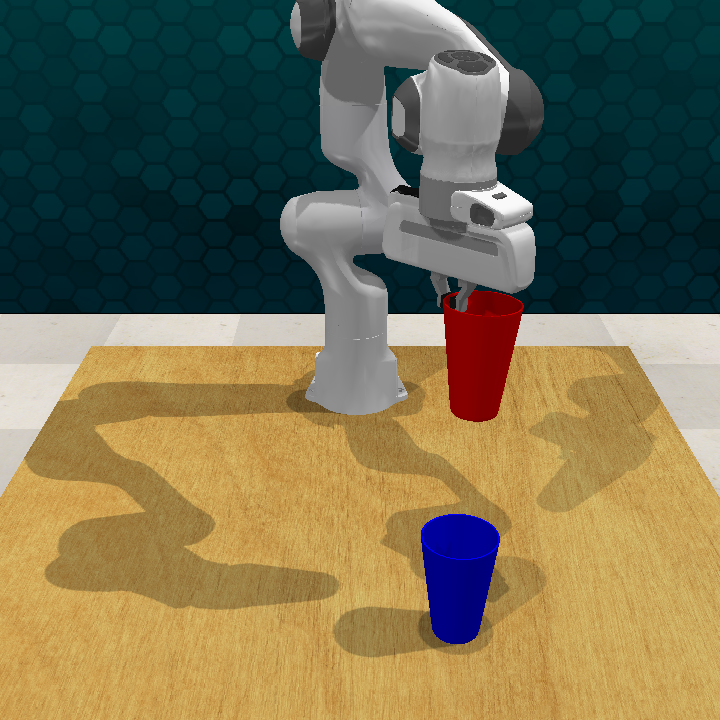} \\
        \raisebox{0.085\linewidth}{Lamp} &
        \includegraphics[width=0.19\linewidth]{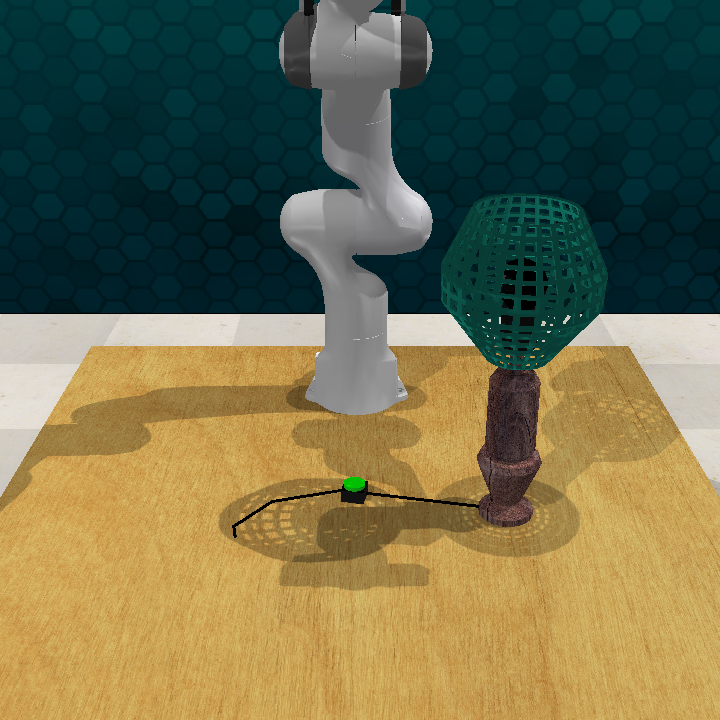} &
        \includegraphics[width=0.19\linewidth]{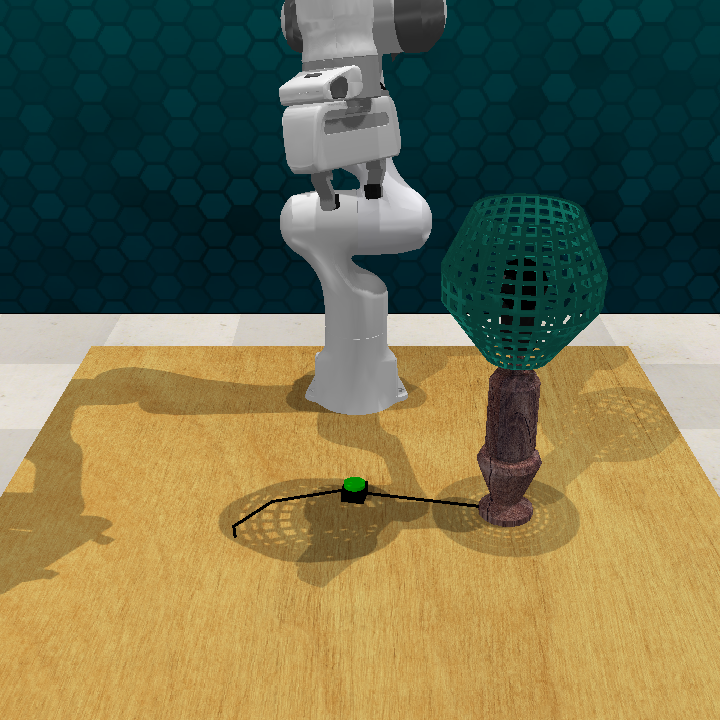} &
        \includegraphics[width=0.19\linewidth]{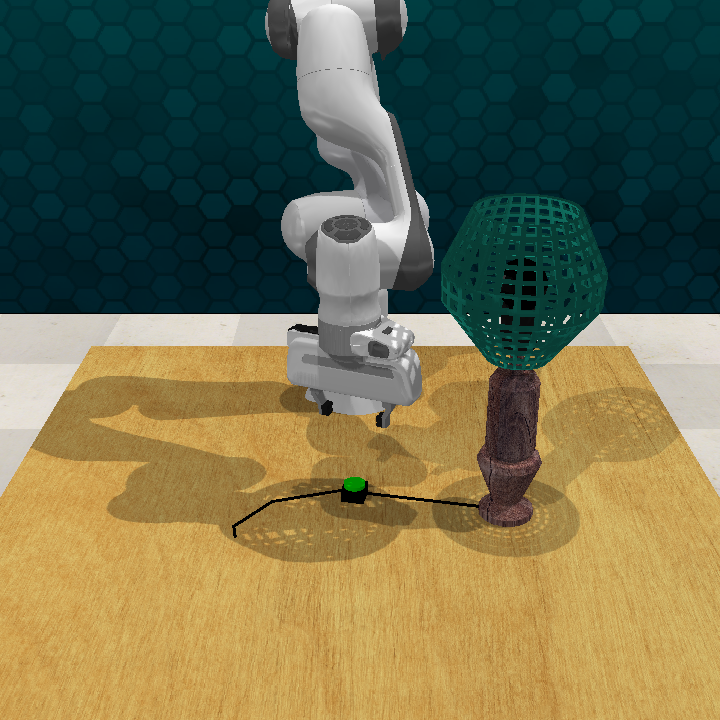} &
        \includegraphics[width=0.19\linewidth]{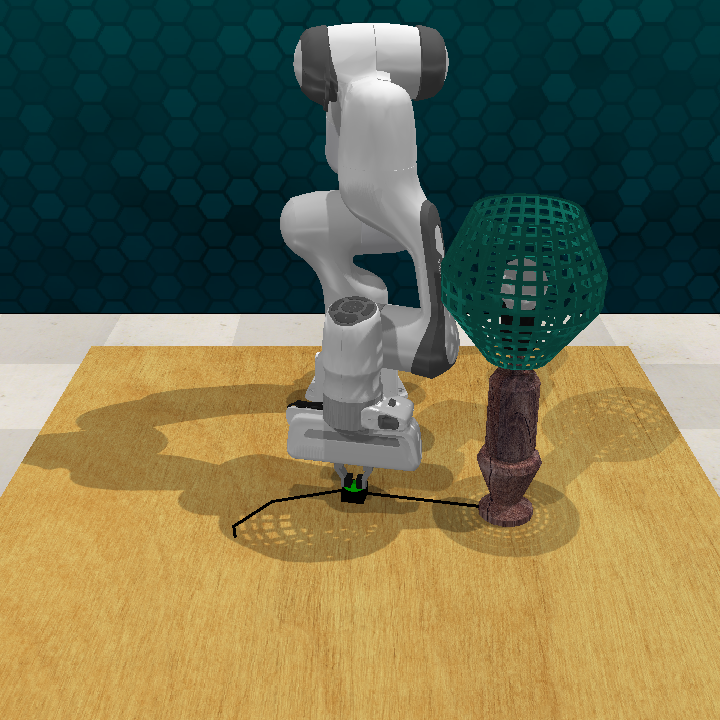} \\
        \bottomrule
    \end{tabular}
    \caption{Simulation visualization examples from three representative manipulation rollouts. Frames in each row are ordered from the beginning to the end of the same episode.}
    \label{fig:appendix-simulation-rollouts}
\end{figure}

Figure~\ref{fig:appendix-simulation-rollouts} shows representative simulation rollouts used for qualitative visualization. The examples cover button pressing, cup manipulation, and lamp operation, and each row presents temporally ordered frames from the same episode.

\begin{figure}[!htbp]
    \centering
    \scriptsize
    \setlength{\tabcolsep}{2pt}
    \renewcommand{\arraystretch}{0.92}
    \begin{tabular}{ccccc}
        \toprule
        Trial & Start & Early & Mid & Final \\
        \midrule
        \raisebox{0.060\linewidth}{Real 1} &
        \includegraphics[width=0.19\linewidth]{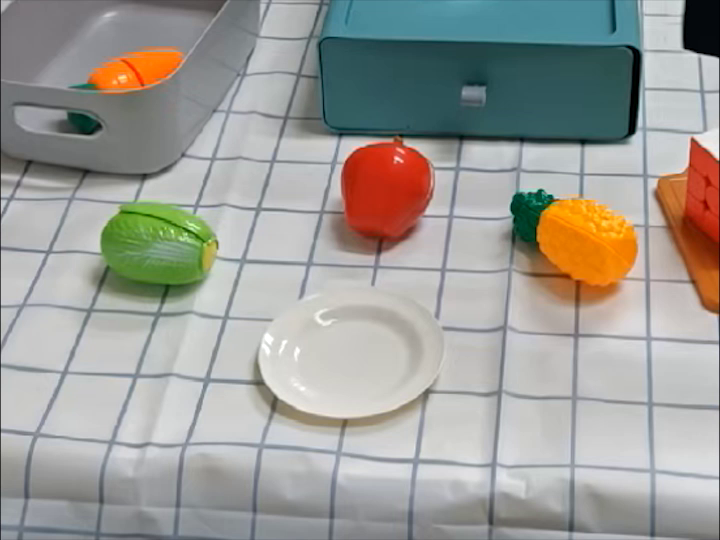} &
        \includegraphics[width=0.19\linewidth]{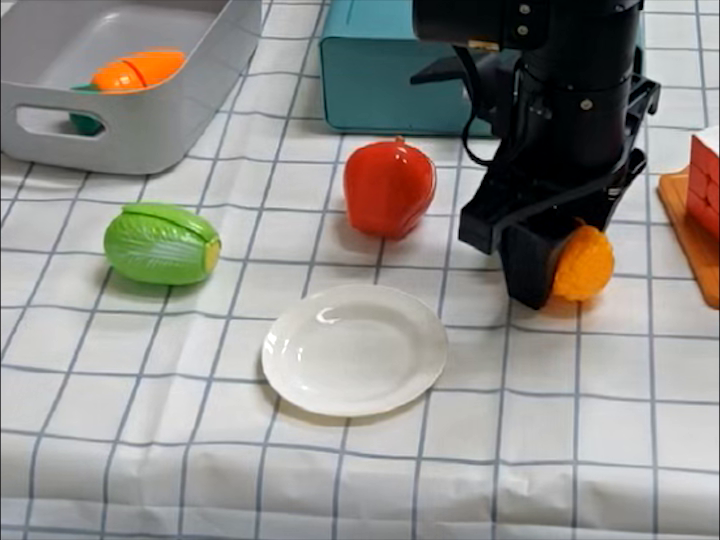} &
        \includegraphics[width=0.19\linewidth]{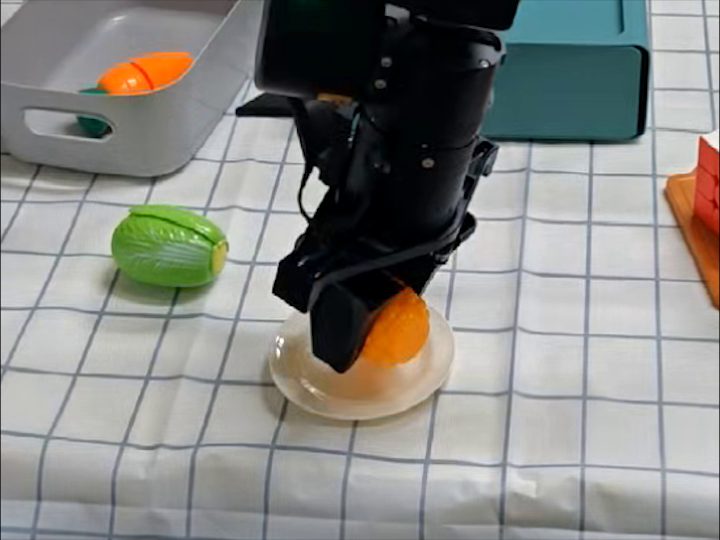} &
        \includegraphics[width=0.19\linewidth]{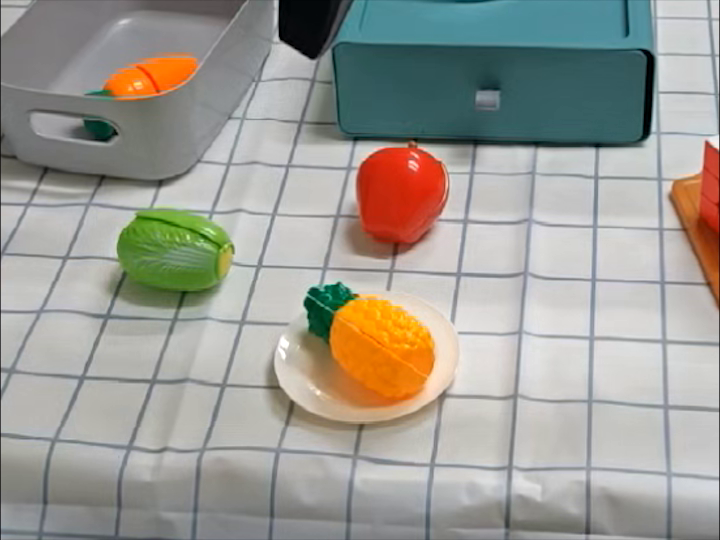} \\
        \raisebox{0.060\linewidth}{Real 2} &
        \includegraphics[width=0.19\linewidth]{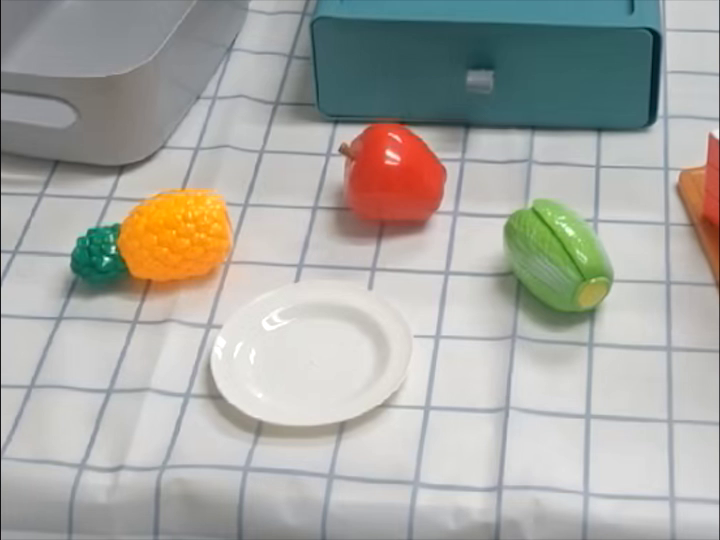} &
        \includegraphics[width=0.19\linewidth]{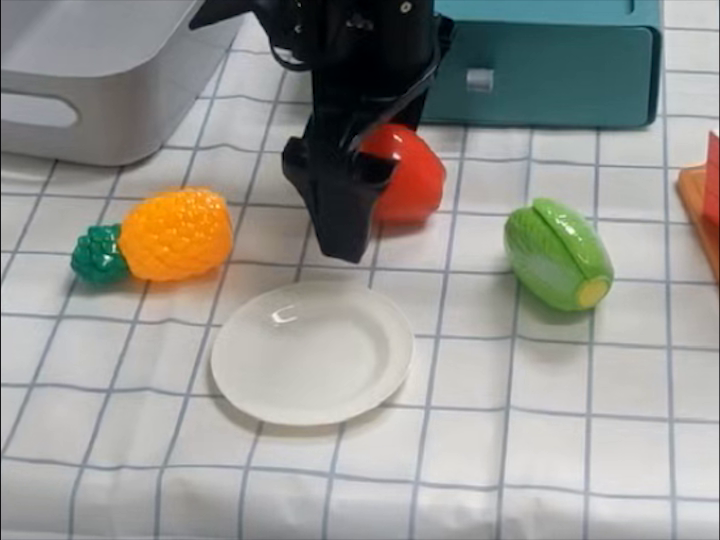} &
        \includegraphics[width=0.19\linewidth]{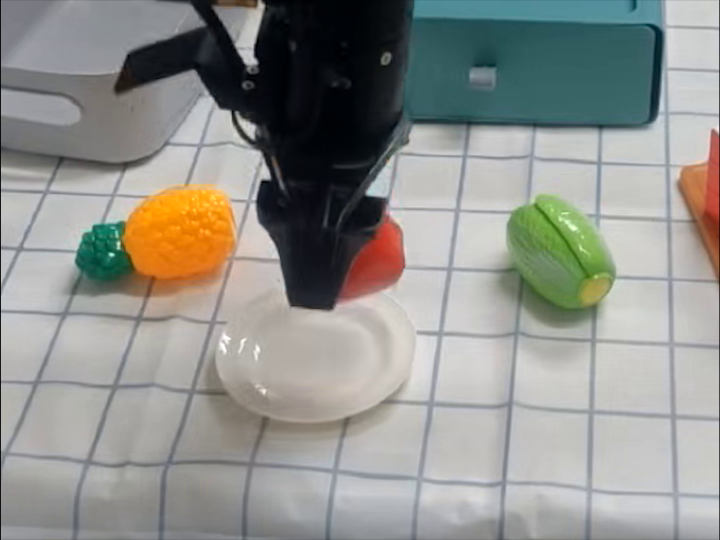} &
        \includegraphics[width=0.19\linewidth]{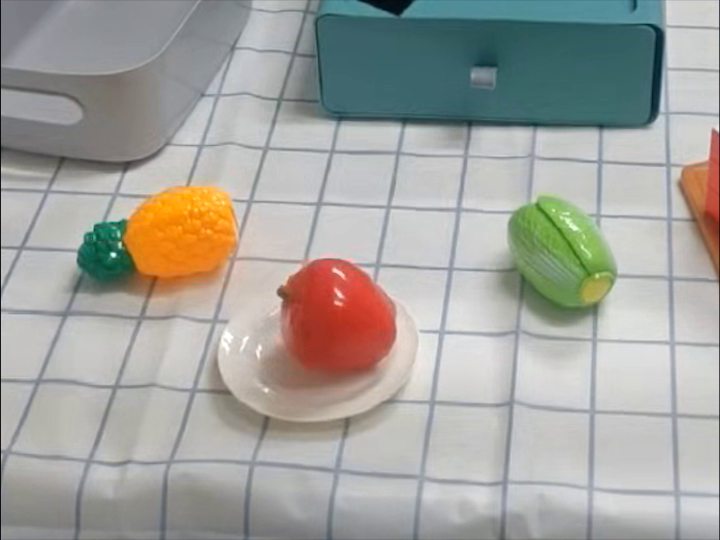} \\
        \raisebox{0.060\linewidth}{Real 3} &
        \includegraphics[width=0.19\linewidth]{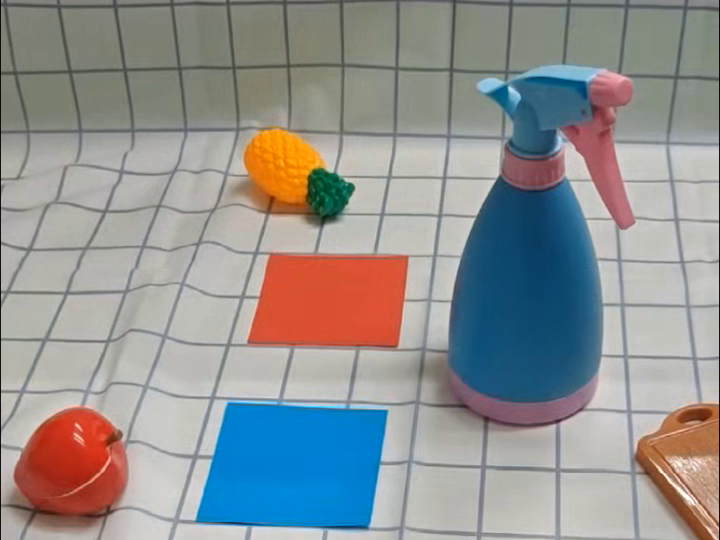} &
        \includegraphics[width=0.19\linewidth]{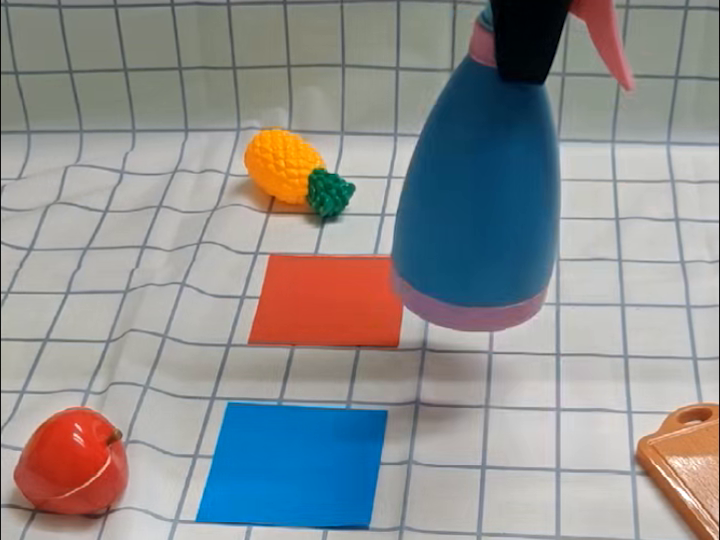} &
        \includegraphics[width=0.19\linewidth]{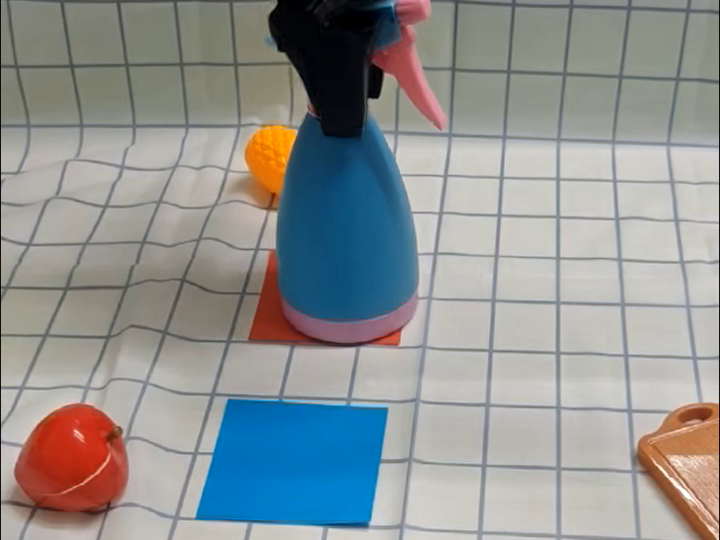} &
        \includegraphics[width=0.19\linewidth]{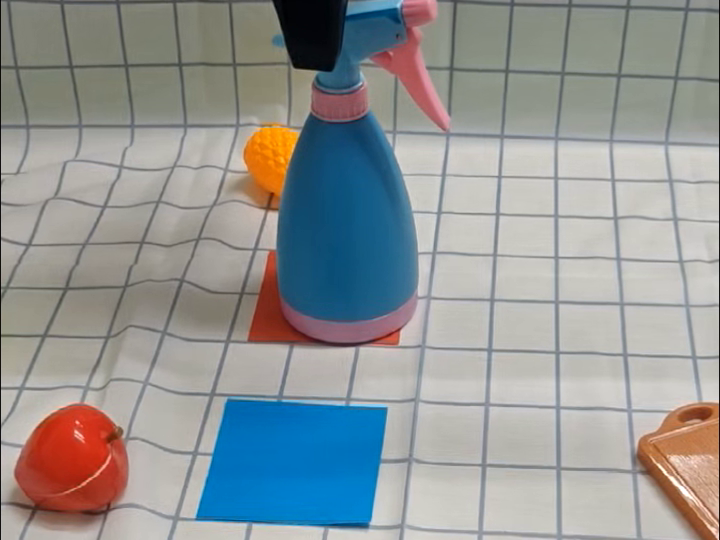} \\
        \bottomrule
    \end{tabular}
    \caption{Additional real-world robot execution visualizations from three recorded trials. Frames in each row are sampled from the same video and ordered from the beginning to the end of the trial.}
    \label{fig:appendix-real-rollouts}
\end{figure}

For the real-world experiments in Section~\ref{sec:realworld-experiments}, the agent explores each task ten times before testing and stores reusable experience in the KnowledgeBank. The model parameters are not updated, so the improvement comes from calibrated 3D perception, retrieval, and planning-time memory use rather than policy fine-tuning. Figure~\ref{fig:appendix-real-rollouts} shows additional real-world execution frames from three recorded robot trials. In open\_drawer, GeneralVLA-2 localizes 3D objects and estimates the drawer orientation, while RoboPoint only localizes targets in the 2D image. In move\_spray\_bottle, the KnowledgeBank learns how high the gripper should lift an object after grasping to avoid collisions, whereas CAP only has simple hand-crafted primitive actions and no dedicated primitive for opening drawers.

\FloatBarrier

\section{Additional Limitations and Failure Modes}
\label{app:limitations-details}

GeoFuse-MV3D inherits several practical constraints from the multi-view reconstruction setting. It assumes that the input masks describe the target object consistently across views and that camera poses are accurate enough for projection-based support checks. Severe mask leakage, missing object regions, or calibration drift can therefore make the soft visual-hull score unreliable. The method also preserves appearance attributes from the trusted source by design, which avoids color drift but limits its ability to repair texture or lighting artifacts.

The governed KnowledgeBank depends on verifier reliability and task coverage. Incorrect verifier judgments may promote weak memories or suppress useful ones, and rare failure modes may remain underrepresented until the robot has observed them. The current real-world evaluation focuses on tabletop manipulation with static objects and short horizons; future work should test longer mobile-manipulation tasks, stronger occlusions, deformable objects, and interactive recovery.
\FloatBarrier

\end{document}